\documentclass{article} 

\usepackage[dvipsnames, table, xcdraw]{xcolor}

\usepackage{utfsym}

\usepackage{iclr2024_conference,times}

\usepackage{booktabs}

\usepackage[hidelinks,pagebackref, breaklinks, colorlinks, linkcolor=BrickRed, citecolor=RoyalBlue]{hyperref}
\usepackage{amsmath, amssymb, mathtools, bm}
\usepackage{multirow, enumitem, subcaption, float, wrapfig}
\usepackage {appendix}
\usepackage{pifont}
\usepackage{wrapfig}
\newcommand{\tablestyle}[2]{\footnotesize \setlength{\tabcolsep}{#1}\renewcommand{\arraystretch}{#2}\centering}

\newcommand{\figref}[1]{Fig.~\ref{#1}}
\newcommand{\tabref}[1]{Tab.~\ref{#1}}
\newcommand{\eqnref}[1]{Eqn.~\ref{#1}}
\newcommand{\secref}[1]{Sec.~\ref{#1}}
\newcommand{\myPara}[1]{\noindent\textbf{#1}}
\newcommand{\sArt}{state-of-the-art~}

\graphicspath{{./images/}}

\usepackage{makecell}
\newcommand{\highlight}[1]{\textbf{\textcolor{BrickRed}{#1}}}

\newcommand{\addFig}[1]{}
\newcommand{\addFigs}[1]{}

\def\eg{\emph{e.g.,~}}
\def\ie{\emph{i.e.,~}}

\newcommand{\nMethod}{DFormer}

\title{\nMethod{}: Rethinking RGBD Representation Learning for Semantic Segmentation}

\author{Bowen Yin$^1$~~~Xuying Zhang$^1$~~~Zhongyu Li$^1$~~~Li Liu$^2$~~~Ming-Ming Cheng$^1$~~~Qibin Hou$^1$\thanks{Qibin Hou is the corresponding author.} \\ \\
$^1$ VCIP, School of Computer Science, Nankai University \\
$^2$ National University of Defense Technology\\
{\tt\small bowenyin@mail.nankai.edu.cn, andrewhoux@gmail.com}
}

\iclrfinalcopy 
\begin{document}

\maketitle

\vspace{-10pt}
\begin{abstract}

We present \nMethod{}, a novel RGB-D pretraining framework to learn transferable representations for RGB-D segmentation tasks.
\nMethod{} has two new key innovations:
1) Unlike previous works that encode RGB-D information with RGB pretrained backbone, we pretrain the backbone using image-depth pairs from ImageNet-1K, and hence the \nMethod{} is endowed with the capacity to encode RGB-D representations; 
2)  \nMethod{} comprises a sequence of RGB-D blocks, which are tailored for encoding both RGB and depth information through a novel building block design.
\nMethod{} avoids the mismatched encoding of the 3D geometry relationships in depth maps by RGB pretrained backbones, which widely lies in existing methods but has not been resolved.
We finetune the pretrained \nMethod{} on two popular RGB-D tasks, \ie RGB-D semantic segmentation and RGB-D salient object detection, with a lightweight decoder head.
Experimental results show that our \nMethod{} achieves new state-of-the-art performance on these two tasks with less than half of the computational cost of the current best methods on two RGB-D semantic segmentation datasets and five RGB-D salient object detection datasets.
%
%
Our code is available at: \href{https://github.com/VCIP-RGBD/DFormer}{https://github.com/VCIP-RGBD/DFormer}.
\end{abstract}

\section{Introduction} \label{sec:intro}

\begin{figure}[!ht]
\centering
\vspace{-0.2cm}
\vspace{-5pt}
\includegraphics[width=0.96\linewidth]{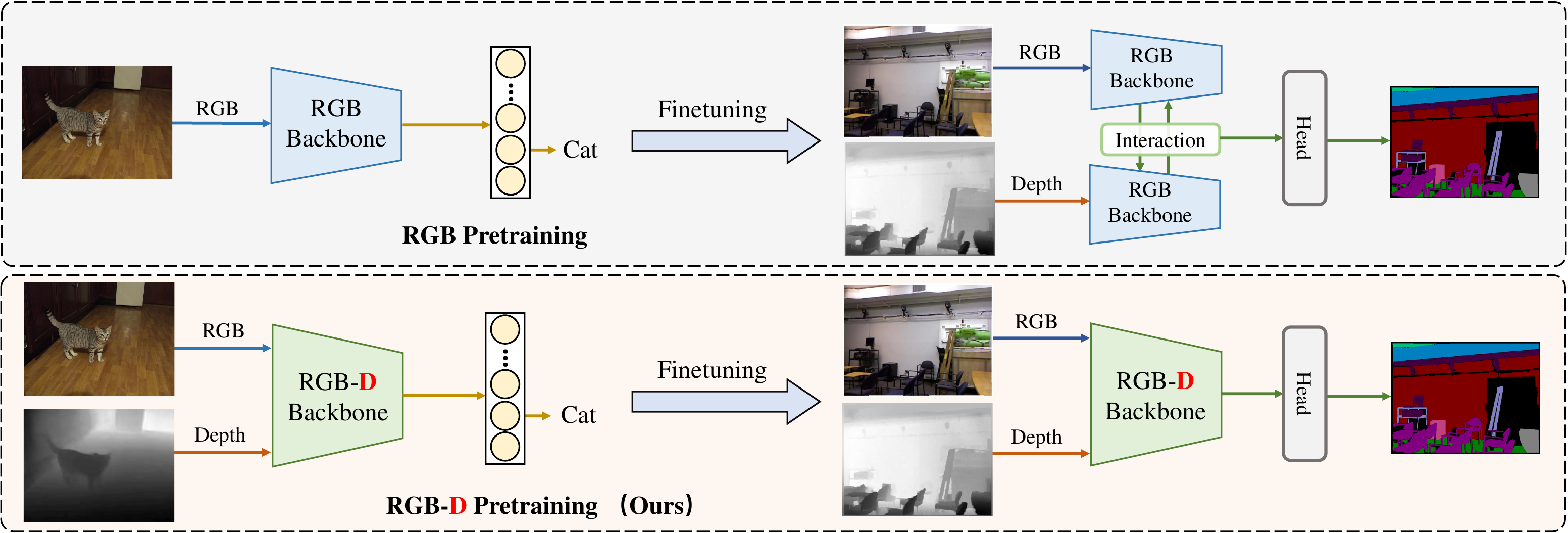}
\vspace{-5pt}
\caption{\small Comparisons between the existing popular training pipeline and ours for RGB-D segmentation. \textbf{RGB pretraining}: Recent mainstream methods adopt two RGB pretrained backbones to separately encode RGB and depth information and fuse them at each stage.
\textbf{RGB-D pretraining}: The RGB-D backbone in \nMethod{} learns transferable RGB-D representations during pretraining and then is finetuned for segmentation.
}\label{fig:compare}
\vspace{-5pt}
\end{figure}

With the widespread use of 3D sensors, RGB-D data is becoming increasingly available to access.
By incorporating 3D geometric information, it would be easier to distinguish instances and context, facilitating the RGB-D research for high-level scene understanding. 
Meanwhile, RGB-D data also presents considerable potential in a large number of applications, \emph{e.g.}, SLAM~\citep{wang2023co}, automatic driving~\citep{huang2022multi}, and robotics~\citep{marchal2020learning}. 
Therefore, RGB-D research has attracted great attention over the past few years.

\figref{fig:compare}~(top) shows the pipeline of current mainstream RGB-D methods. 
As can be observed, the features of the RGB images and depth maps are respectively extracted from two individual RGB pretrained backbones.
The interactions between the information of these two modalities are performed during this process.
Although the existing methods~\citep{wang2022multimodal,zhang2023delivering} have achieved excellent performance on several benchmark datasets, there are three issues that cannot be ignored: 
i) 
The backbones in the RGB-D tasks take an image-depth pair as input, 
which is inconsistent with the input of an image in RGB pretraining, 
causing a huge representation distribution shift;
ii) 
The interactions are densely performed between the RGB branch and depth branch during finetuning, which may destroy the representation distribution within the pretrained RGB backbones;
iii) The dual backbones in RGB-D networks bring more computational cost compared to standard RGB methods, which is not efficient.
We argue that an important reason leading to these issues is the pretraining manner.
The depth information is not considered during pretraining.

\begin{wrapfigure}{r}{0.5\textwidth}
\vspace{-5.5pt}
\centering
\includegraphics[width=0.99\linewidth]{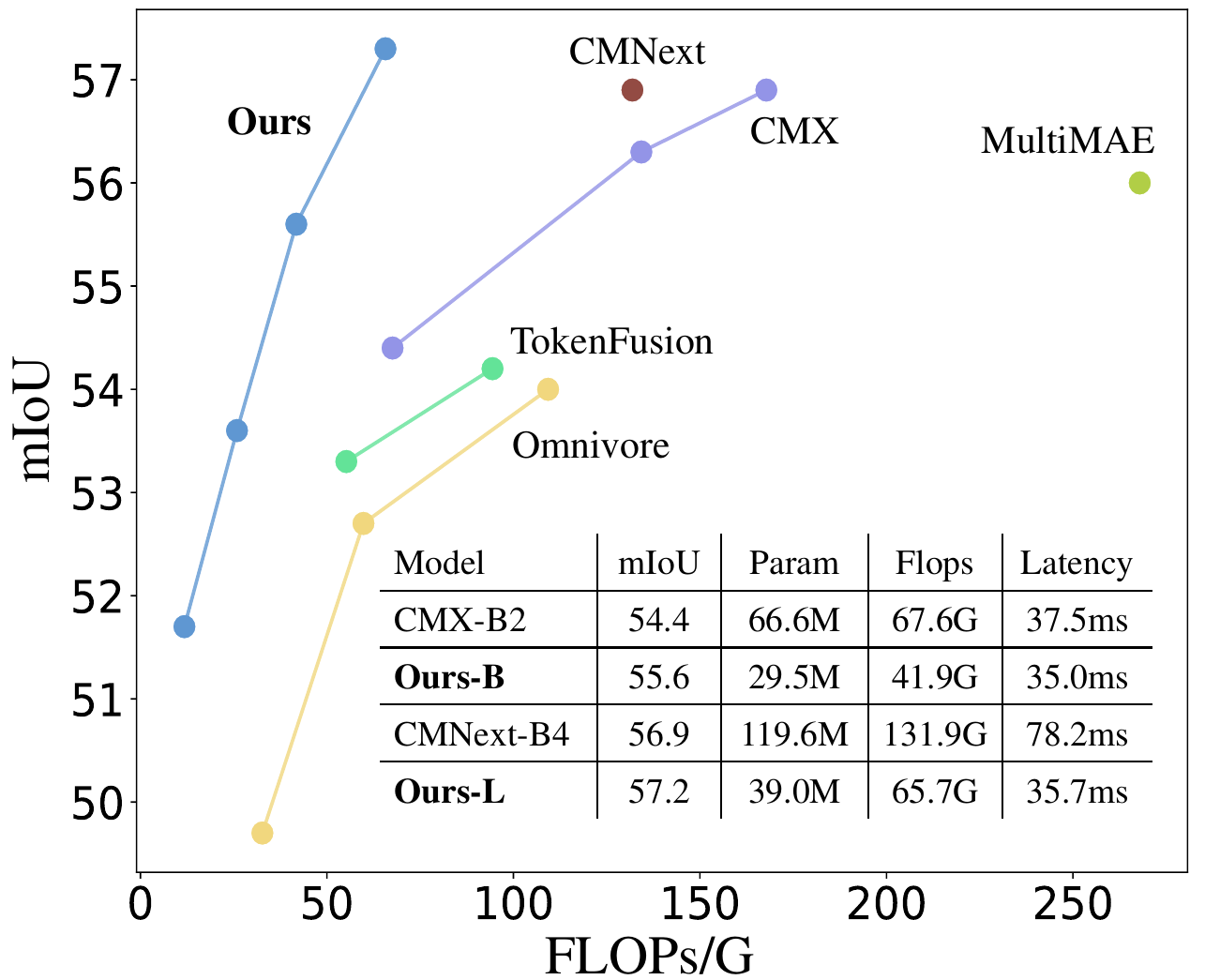}
\vspace{-20pt}
\caption{\small Performance \emph{vs.} computational cost on the NYUDepthv2 dataset~\citep{silberman2012nyu_dataset}. \nMethod{} achieves the state-of-the-art 57.2\% mIoU and the best trade-off compared to other methods.}\label{fig:trade-off}
\vspace{-10pt}
\end{wrapfigure}

Taking the above issues into account, a straightforward question arises: Is it possible to specifically design an RGB-D pretraining framework to eliminate this gap?
This motivates us to present a novel RGB-D pretraining framework, termed \nMethod{}, as shown in \figref{fig:compare}~(bottom).
During pretraining, we consider taking image-depth pairs
\footnote{For depth maps, we employ a widely used depth estimation model~\citep{bhat2021adabins} to predict depth map for each RGB image,which we found works well.}, not just RGB images, as input and propose to build interactions between RGB and depth features within the building blocks of the encoder.
Therefore, the inconsistency between the inputs of pretraining and finetuning can be naturally avoided.
In addition, during pretraining, the RGB and depth features can efficiently interact with each other in each building block, avoiding the heavy interaction modules outside the backbones, which is mostly adopted in
current dominant methods.
Furthermore, we also observe that the depth information only needs a small portion of channels to encode.
There is no need to use a whole RGB pretrained backbone to extract depth features as done in previous works.
As the interaction starts from the pretraining stage, the interaction efficiency can be largely improved compared to previous works
as shown in \figref{fig:trade-off}.
%

We demonstrate the effectiveness of \nMethod{} on two popular RGB-D downstream tasks, \emph{i.e.}, semantic segmentation and salient object detection.
By adding a lightweight decoder on top of our pretrained RGB-D backbone, \nMethod{} sets new \sArt records with less computation costs compared to previous methods. 
Remarkably, our largest model, \nMethod{}-L, achieves a new \sArt result, \ie 57.2\% mIoU on NYU Depthv2 with less than half of the computations of the second-best method CMNext~\citep{zhang2023delivering}.
Meanwhile, our lightweight model \nMethod{}-T is able to achieve 51.8\% mIoU on NYU Depthv2 with only 6.0M parameters and 11.8G Flops. 
Compared to other recent models, our approach achieves the best trade-off between segmentation performance and computations.

To sum up, our main contributions can be summarized as follows:
\begin{itemize}
    %
    \item We present a novel RGB-D pretraining framework, termed \nMethod{},
    with a new interaction method to fuse the RGB and depth features to provide transferable representations for RGB-D downstream tasks.
    \item We find that in our framework it is enough to use a small portion of channels to encode the depth information compared to RGB features, an effective way to reduce model size.
    \item Our \nMethod{} achieves new \sArt performance with less than half of computational cost of the current best methods on two RGB-D segmentation datasets and five RGB-D salient object detection datasets.
\end{itemize}

\section{Proposed \nMethod{}}



\begin{figure}[tp]
\centering
\includegraphics[width=\linewidth]{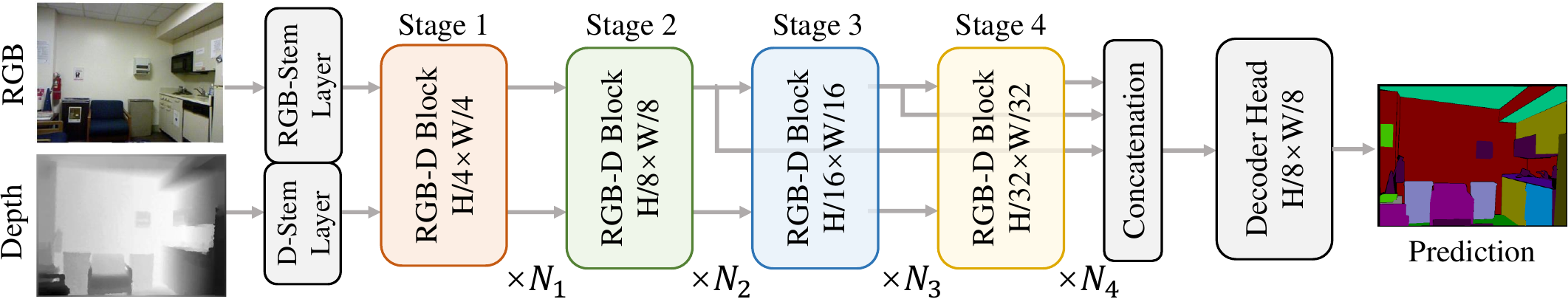}
\vspace{-15pt}
\caption{\small Overall architecture of the proposed \nMethod{}. 
First, we use the pretrained \nMethod{} to encode the RGB-D data. Then, the features from the last three stages are concatenated and delivered to a lightweight decoder head for final prediction. Note that only the RGB features from the encoder are used in the decoder.
}\label{fig:architecture}
\vspace{-10pt}
\end{figure}

\figref{fig:architecture} illustrates the overall architecture of our \nMethod{}, which follows the popular encoder-decoder framework.
In particular, the hierarchical encoder is designed to generate high-resolution coarse features and low-resolution fine features, and a lightweight decoder is employed to transform these visual features into task-specific predictions.

Given an RGB image and the corresponding depth map with spatial size of $H \times W$, they are first separately processed by two parallel stem layers consisting of two convolutions with kernel size $3 \times 3$ and stride 2. 
Then, the RGB features and depth features are fed into the hierarchical encoder to encode multi-scale features
at $\{ 1/4, 1/8, 1/16, 1/32 \}$ of the original image resolution.
Next, we pretrain this encoder using the image-depth pairs from ImageNet-1K using the classification objective to generate the transferable RGB-D representations.
Finally, we send the visual features from the pretrained RGB-D encoder to the decoder to produce predictions, \emph{e.g.}, segmentation maps with a spatial size $H \times W$.
In the rest of this section, we will describe the encoder, RGB-D pretraining framework, and task-specific decoder in detail.

\subsection{Hierarchical Encoder}

As shown in \figref{fig:architecture}, our hierarchical encoder is composed of four stages, which are utilized to generate multi-scale RGB-D features.
Each stage contains a stack of RGB-D blocks. 
Two convolutions with kernel size $3 \times 3$ and stride 2 are used to down-sample RGB and depth features, respectively, between two consecutive stages.

\begin{wrapfigure}{r}{0.5\textwidth}
  \setlength{\abovecaptionskip}{2pt}
  \vspace{-10pt}
  \centering
  \includegraphics[width=0.95\linewidth]{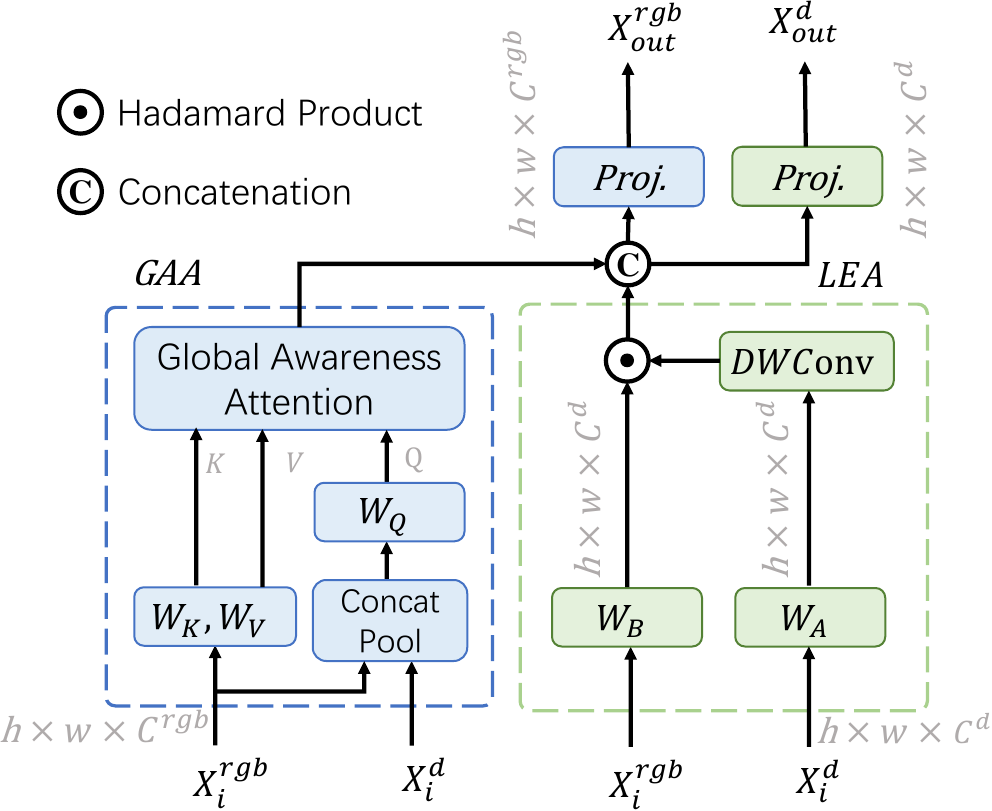}
  \caption{\small Diagrammatic details on how to conduct interactions between RGB and depth features. 
  }
  \vspace{-15pt}
  \label{fig:block}
\end{wrapfigure}

\myPara{Building Block.}
Our building block is mainly composed of the global awareness attention (GAA) module and the local enhancement attention (LEA) module and builds interaction between the RGB and depth modalities. 
GAA incorporates depth information and aims to enhance the capability of object localization from a global perspective, while LEA adopts a large-kernel convolution to capture the local clues from the depth features, which can refine the details of the RGB representations.  
%
%
The details of the interaction modules are shown in \figref{fig:block}.
Our GAA fuses depth and RGB features to build relationships across the whole scene, enhancing 3D awareness and further helping capture semantic objects. 
Different from the self-attention mechanism~\citep{vaswani2017attention} that introduces quadratic computation growth as the pixels or tokens increase, the Query $(Q)$ in GAA is down-sampled to a fixed size and hence the computational complexity can be reduced.  
\tabref{tab:gaa} illustrates that fusing depth features with $Q$ is adequate and there is no need to combine them with $K$ or $V$, which brings computation increment but no performance improvement.
%
So, $Q$ comes from the concatenation of the RGB features and depth features, while key ($K$) and value ($V$) are extracted from RGB features.
Given the RGB features $X^{rgb}_{i}$ and depth features $X^{d}_{i}$, the above process can be formulated as:
\begin{equation}
\begin{aligned}
   Q = \mathrm{Linear}(\mathrm{Pool}_{k\times k}([X^{rgb}_{i},X^{d}_{i}])),
      K = \mathrm{Linear}(X^{rgb}_{i}), 
   V = \mathrm{Linear}(X^{rgb}_{i}),
  \label{eq:Q1}
\end{aligned}
\end{equation}
where $[\cdot, \cdot]$ denotes the concatenation operation along the channel dimension, $\mathrm{Pool}_{k\times k}(\cdot)$ performs adaptively average pooling operation across the spatial dimensions to $k\times k$ size, and $\mathrm{Linear}$ is linear transformation.
%
Based on the generated $Q \in \mathbb{R}^{k\times k\times C^{d}}$, $K \in \mathbb{R}^{ h\times w\times C^{d}}$, and $V \in \mathbb{R}^{h\times w\times C^{d}}$, where $h$ and $w$ are the height and width of features in the current stage, we formulate the GAA as follows:
\begin{equation} 
   X_{GAA} = \mathrm{UP}(V\cdot \mathrm{Softmax} (\frac{{Q}^{\top}{K}}{\sqrt{C^{d}}})),
  \label{eq:GAA1}
\end{equation}
where $\mathrm{UP}(\cdot)$ is a bilinear upsampling operation that converts the spatial size from $k\times k$ to $h\times w$.
In practical use, \eqnref{eq:GAA1} can also be extended to a multi-head
version, as done in the original self-attention~\citep{vaswani2017attention}, 
to augment the feature representations.

We also design the LEA module to capture more local details, which can be regarded as a supplement to the GAA module.
%
Unlike most previous works that use addition and concatenation to fuse the RGB features and depth features.
We conduct a depth-wise convolution with a large kernel on the depth features and use the resulting features as attention weights to reweigh the RGB features via a simple Hadamard product inspired by ~\citep{hou2022conv2former}.
This is reasonable in that adjacent pixels with similar depth values often belong to the same object and the 3D geometry information thereby can be easily embedded into the RGB features.
%
To be specific, the calculation process of LEA can be defined as follows:
\begin{equation} 
   X_{LEA} = \mathrm{DConv}_{k \times k}(\mathrm{Linear}(X^{d}_{i})) \odot \mathrm{Linear}(X^{rgb}_{i}), 
  \label{eq:A}
\end{equation}
where DConv$_{k\times k}$ is a depth-wise convolution with kernel size $k\times k$ and $\odot$ is the Hadamard product.

To preserve the diverse appearance information, we also build a base module to transform the RGB features $X_{i}^{rgb}$ to $X_{Base}$, which has the same spatial size as $X_{GAA}$ and $X_{LEA}$. 
The calculation process of $X_{Base}$ can be defined as follows:
\begin{equation} 
   X_{Base} = \mathrm{DConv}_{k\times k}(\mathrm{Linear}(X^{rgb}_{i})) \odot \mathrm{Linear}(X^{rgb}_{i}).
  \label{eq:base}
\end{equation}
Finally, the features, \ie $X_{GAA} \in \mathbb{R}^{h_{i} \times w_{i} \times C_{i}^{d}}$, $X_{LEA} \in \mathbb{R}^{h_{i} \times w_{i} \times C_{i}^{d}}$, $X_{Base} \in \mathbb{R}^{h_{i} \times w_{i} \times C_{i}^{rgb}}$, are fused together by concatenation and linear projection to update the RGB features $X_{out}^{rgb}$ and depth features $X_{out}^{d}$.




\myPara{Overall Architecture.}
We empirically observe that encoding depth features requires fewer parameters compared to the RGB ones due to their less semantic information, which is verified in \figref{fig:ratio} and illustrated in detail in the experimental part.
To reduce model complexity in our RGB-D block, we use a small portion of channels to encode the depth information.
%
%
Based on the configurations of the RGB-D blocks in each stage, we design a series of \nMethod{} encoder variants, termed \nMethod{}-T, \nMethod{}-S, \nMethod{}-B, and \nMethod{}-L, respectively, with the same architecture but different model sizes. 
\nMethod{}-T is a lightweight encoder for fast inference, while \nMethod{}-L is the largest one for attaining better performance.
For detailed configurations, readers can refer to \tabref{tab:conf}.

\subsection{RGB-D pretraining}\label{sec:RGBDPretrain}
The purpose of RGB-D pretraining is to endow the backbone with the ability to achieve the interaction between RGB and depth modalities and generate transferable representations with rich semantic and spatial information.
To this end, we first apply a depth estimator, \emph{e.g.}, Adabin~\citep{bhat2021adabins},  on the ImageNet-1K dataset~\citep{russakovsky2015imagenet} to generate a large number of image-depth pairs.
%
%
Then, we add a classifier head on the top of the RGB-D encoder to build the classification network for pretraining.
Particularly, the RGB features from the last stage are flattened along the spatial dimension and fed into the classifier head.
The standard cross-entropy loss is employed as our optimization objective, and the network is pretrained on RGB-D data for 300 epochs, like ConvNext~\citep{liu2022convnet}.
%
%
%
Following previous works~\citep{liu2022convnet,guo2022visual}, the AdamW ~\citep{loshchilov2017decoupled}  with learning rate 1e-3 and weight decay 5e-2 is employed as our optimizer, and the batch size is set to 1024.
%
%
More specific settings for each variant of \nMethod{} are described in the appendix.

\subsection{Task-specific Decoder.}
For the applications of our \nMethod{} to downstream tasks, we just add a lightweight decoder on top of the pretrained RGBD backbone to build the task-specific network.
After being finetuned on corresponding benchmark datasets, the task-specific network is able to generate great predictions, without using extra designs like fusion modules~\citep{chen2020sa_gate,zhang2022cmx}. 

Take RGB-D semantic segmentation as an example.
Following SegNext~\citep{guo2022segnext}, we adopt a lightweight Hamburger head~\citep{geng2021attention} to aggregate the multi-scale RGB features from the last three stages of our pretrained encoder. 
%
Note that, our decoder only uses the $X^{rgb}$ features, while other methods~\citep{zhang2022cmx,wang2022multimodal,zhang2023delivering} mostly design modules that fuse both modalities features $X^{rgb}$ and $X^{d}$ for final predictions.
We will show in our experiments that our $X^{rgb}$ features can efficiently extract the 3D geometry clues from the depth modality thanks to our powerful RGB-D pretrained encoder. 
Delivering the depth features $X^{d}$ to the decoder is not necessary.

\section{Experiments}\label{sec:experiment}
\begin{table}[tp]
  \centering
    \setlength{\tabcolsep}{4pt}
    \scriptsize
    \caption{\small Results on NYU Depth V2~\citep{silberman2012nyu_dataset} and SUN-RGBD~\citep{song2015sun_rgbd}.
    Some methods do not report the results or settings on the SUN-RGBD datasets, so we reproduce them with the same training config. $^{\dag}$ indicates our implemented results. All the backbones are pre-trained on ImageNet-1K.}\label{tab:rgbd_sota}
    \vspace{-10pt}
    \centering
    \renewcommand{\arraystretch}{0.9}
    	\begin{tabular}{lccccccccc}
        \toprule
        \textbf{\multirow{2}{*}{Model}}  &\textbf{\multirow{2}{*}{Backbone}}   & \textbf{\multirow{2}{*}{Params}} &\multicolumn{3}{c}{NYUDepthv2} & \multicolumn{3}{c}{SUN-RGBD} &\textbf{\multirow{2}{*}{Code}}\\ \cmidrule(lr){4-6}\cmidrule(lr){7-9}
         && &\textbf{Input size}& \textbf{Flops}& \textbf{mIoU}   &\textbf{Input size}& \textbf{Flops}& \textbf{mIoU} & \\
        \midrule\midrule
        ACNet$_{\rm 19}$~(\citeauthor{hu2019acnet})&ResNet-50&116.6M&$480\times 640$ &126.7G& 48.3 &$530\times 730$&163.9G&48.1&\href{https://github.com/anheidelonghu/ACNet}{Link} \\
        SGNet$_{\rm 20}$~(\citeauthor{chen2021spatial_guided})&ResNet-101&64.7M&$480\times 640$ &108.5G& 51.1 &$530\times 730$&151.5G&48.6&\href{https://github.com/LinZhuoChen/SGNet}{Link}\\
        SA-Gate$_{\rm 20}$~(\citeauthor{chen2020sa_gate})&ResNet-101&110.9M&$480\times 640$ &193.7G& 52.4 &$530\times 730$&250.1G&49.4&\href{https://github.com/charlesCXK/RGBD_Semantic_Segmentation_PyTorch}{Link} \\
        CEN$_{\rm 20}$~(\citeauthor{wang2020deep})&ResNet-101&118.2M&$480\times 640$&618.7G&51.7&$530\times 730$&790.3G&50.2&\href{https://github.com/yikaiw/CEN}{Link}\\
        CEN$_{\rm 20}$~(\citeauthor{wang2020deep})&ResNet-152&133.9M&$480\times 640$&664.4G&52.5&$530\times 730$&849.7G&51.1&\href{https://github.com/yikaiw/CEN}{Link}\\
        ShapeConv$_{\rm 21}$~(\citeauthor{cao2021shapeconv})&ResNext-101&86.8M&$480\times 640$ &124.6G& 51.3 &$530\times 730$&161.8G&48.6&\href{https://github.com/hanchaoleng/ShapeConv}{Link} \\ 
        ESANet$_{\rm 21}$~(\citeauthor{seichter2021efficient})&ResNet-34&31.2M&$480\times 640$&34.9G&50.3&$480\times 640$&34.9G&48.2&\href{https://github.com/TUI-NICR/ESANet}{Link}\\
        FRNet$_{\rm 22}$~(\citeauthor{zhou2022frnet})&ResNet-34&85.5M&$480\times 640$&115.6G&53.6&$530\times 730$&150.0G&51.8&\href{https://github.com/EnquanYang2022/FRNet}{Link}\\
        PGDENet$_{\rm 22}$~(\citeauthor{zhou2022pgdenet})&ResNet-34&100.7M&$480\times 640$&178.8G&53.7&$530\times730$&229.1G&51.0&\href{https://github.com/EnquanYang2022/PGDENet}{Link}\\
        EMSANet$_{\rm 22}$~(\citeauthor{seichter2022efficient})&ResNet-34&46.9M&$480\times 640$&45.4G&51.0&$530\times 730$&58.6G&48.4&\href{https://github.com/TUI-NICR/EMSANet}{Link}\\ 
        TokenFusion$_{\rm 22}$~(\citeauthor{wang2022multimodal})& MiT-B2&26.0M&$480\times640$&55.2G&53.3&$530\times 730$&71.1G&50.3$^{\dag}$&\href{https://github.com/yikaiw/TokenFusion}{Link}\\
        TokenFusion$_{\rm 22}$~(\citeauthor{wang2022multimodal}) &MiT-B3&45.9M&$480\times640$&94.4G&54.2&$530\times 730$&122.1G&51.0$^{\dag}$&\href{https://github.com/yikaiw/TokenFusion}{Link}\\
        MultiMAE$_{\rm 22}$~(\citeauthor{bachmann2022multimae})&ViT-B&95.2M&$640\times640$&267.9G&56.0&$640\times 640$&267.9G&51.1$^{\dag}$&\href{https://github.com/EPFL-VILAB/MultiMAE}{Link}\\
        Omnivore$_{\rm 22}$~(\citeauthor{girdhar2022omnivore})&Swin-T&29.1M&$480\times 640$&32.7G&49.7&$530\times730$&---&---&\href{https://github.com/facebookresearch/omnivore}{Link}\\
        Omnivore$_{\rm 22}$~(\citeauthor{girdhar2022omnivore})&Swin-S&51.3M&$480\times 640$&59.8G&52.7&$530\times730$&---&---&\href{https://github.com/facebookresearch/omnivore}{Link}\\
        Omnivore$_{\rm 22}$~(\citeauthor{girdhar2022omnivore})&Swin-B&95.7M&$480\times 640$&109.3G&54.0&$530\times730$&---&---&\href{https://github.com/facebookresearch/omnivore}{Link}\\
        CMX$_{\rm 22}$~(\citeauthor{zhang2022cmx})&MiT-B2&66.6M&$480\times 640$&67.6G&54.4&$530\times730$&86.3G&49.7&\href{https://github.com/huaaaliu/RGBX_Semantic_Segmentation}{Link}\\
        CMX$_{\rm 22}$~(\citeauthor{zhang2022cmx})&MiT-B4&139.9M&$480\times 640$&134.3G&56.3&$530\times730$&173.8G&52.1&\href{https://github.com/huaaaliu/RGBX_Semantic_Segmentation}{Link}\\
        CMX$_{\rm 22}$~(\citeauthor{zhang2022cmx})&MiT-B5&181.1M&$480\times 640$&167.8G&56.9&$530\times730$&217.6G&52.4&\href{https://github.com/huaaaliu/RGBX_Semantic_Segmentation}{Link}\\
        CMNext$_{\rm 23}$~(\citeauthor{zhang2023delivering})&MiT-B4&119.6M&$480\times640$&131.9G&56.9&$530\times730$&170.3G&51.9$^{\dag}$&\href{https://github.com/jamycheung/DELIVER}{Link}\\
        \midrule
        \rowcolor{gray!15}\nMethod{}-T &Ours-T&6.0M&$480\times 640$& 11.8G&51.8&$530\times730$&15.1G&48.8&\href{https://github.com/VCIP-RGBD/DFormer}{Link}\\
        \rowcolor{gray!15}\nMethod{}-S &Ours-S&18.7M&$480\times 640$ &25.6G&53.6 &$530\times730$&33.0G&50.0&\href{https://github.com/VCIP-RGBD/DFormer}{Link}\\
        \rowcolor{gray!15} \nMethod{}-B&Ours-B&29.5M&$480\times 640$ &41.9G&55.6&$530\times730$&54.1G&51.2&\href{https://github.com/VCIP-RGBD/DFormer}{Link}\\
        \rowcolor{gray!15}\nMethod{}-L&Ours-L&39.0M&$480\times 640$ &65.7G&\textbf{\highlight{57.2}}&$530\times730$&83.3G&\highlight{52.5}&\href{https://github.com/VCIP-RGBD/DFormer}{Link} \\
        \bottomrule
        \end{tabular}
    
    \hspace{\fill}
    \hspace{\fill}
    \vspace{-10pt}
\end{table}

\subsection{RGB-D Semantic Segmentation}\label{sec:exp_seg}

\myPara{Datasets\& implementation details.}
Following the common experiment settings of RGB-D semantic segmentation methods~\citep{xie2021segformer,guo2022segnext}, we finetune and evaluate the \nMethod{} on two widely used datasets, \ie NYUDepthv2~\citep{silberman2012nyu_dataset} and SUN-RGBD~\citep{song2015sun_rgbd}.
Following SegNext~\cite{guo2022segnext}, we employ Hamburger~\citep{geng2021attention}, a lightweight head, as the decoder to build our RGB-D semantic segmentation network.
During finetuning, we only adopt two common data augmentation strategies, \ie random horizontal flipping and random scaling (from 0.5 to 1.75).
The training images are cropped and resized to $480\times 640$ and $480\times 480$ respectively for NYU Depthv2 and SUN-RGBD benchmarks. 
Cross-entropy loss is utilized as the optimization objective.
We use AdamW~\citep{kingma2014adam} as our optimizer with an initial learning rate of 6e-5 and the poly decay schedule. 
Weight decay is set to 1e-2.
During testing, we employ mean Intersection over Union (mIoU), which is averaged across semantic categories, as the primary evaluation metric to measure the segmentation performance. 
Following recent works~\citep{zhang2022cmx,wang2022multimodal,zhang2023delivering}, we adopt multi-scale (MS) flip inference strategies with scales $\left \{0.5, 0.75, 1, 1.25, 1.5\right \}$.
%



\myPara{Comparison with state-of-the-art methods.}
We compare our \nMethod{} with 13 recent RGB-D semantic segmentation methods on the NYUDepthv2~\citep{silberman2012nyu_dataset} and SUN-RGBD~\citep{song2015sun_rgbd} datasets. 
These methods are chosen according
to three criteria: a) recently published, b) representative,
and c) with open-source code.
As shown in \tabref{tab:rgbd_sota}, our \nMethod{} achieves new \sArt performance across these two benchmark datasets. 
We also plot the performance-efficiency curves of different methods on the validation set of the NYUDepthv2~\citep{silberman2012nyu_dataset} dataset in \figref{fig:trade-off}.
It is clear that \nMethod{} achieves much better performance and computation trade-off compared to other methods.
Particularly, \nMethod{}-L yields 57.2\% mIoU with 39.0M parameters and 65.7G Flops, while the recent best RGB-D semantic segmentation method, \ie CMX (MiT-B2), only achieves 54.4\% mIoU using 66.6M parameters and 67.6G Flops.
It is noteworthy that our \nMethod{}-B can outperform CMX (MIT-B2) by 1.2\% mIoU with half of the parameters (29.5M, 41.9G vs 66.6M, 67.6G).
Moreover, the qualitative comparisons between the semantic segmentation results of our \nMethod{} and CMNext~\citep{zhang2023delivering} in \figref{fig:more_vis} of the appendix further demonstrate the advantage of our method.
In addition, the experiments on SUN-RGBD~\citep{song2015sun_rgbd} also present similar advantages of our \nMethod{} over other methods.
%
%
These consistent improvements indicate that our RGB-D backbone can more efficiently build interaction between RGB and depth features, and hence yields better performance with even lower computational cost.

\begin{table*}[t]
\scriptsize
\setlength\tabcolsep{1.6pt}
\setlength\extrarowheight{1pt}
\begin{center}
\caption{\footnotesize Quantitative comparisons on RGB-D SOD benchmarks. The best results are \highlight{highlighted}.
}
\vspace{-10pt}
\label{tab:sod}
\resizebox{0.99\linewidth}{!}{
\begin{tabular*}{1.09\linewidth}{lcc|| llll|llll|llll|llll|llll}
\toprule
 Dataset & Param  & Flops &\multicolumn{4}{c}{DES(135)}  & \multicolumn{4}{c}{NLPR(300)}   & \multicolumn{4}{c}{NJU2K(500)}   & \multicolumn{4}{c}{STERE(1,000)}  & \multicolumn{4}{c}{SIP(929)}   \\
\cline{4-7} \cline{8-11} \cline{12-15} \cline{16-19} \cline{20-23} 

Metric & \makecell{(M)} &\makecell{(G)}& $M$ &  $F$  &  $S$ &  $E$  &  $M$ &  $F $  &  $S$ &  $E$  &  $M$ &  $F $  &  $S$ &  $E$  &  $M$ &  $F $  &  $S$ &  $E$  &  $M$ &  $F $  &  $S$ &  $E$  \\
\hline

BBSNet$_{21}$ (\citeauthor{zhai2021bifurcated})& 49.8& 31.3& .021 & .942 & .934  &.955 & .023 & .927 & .930  &.953 & .035 & .931 & .920  &.941 & .041 & .919 & .908  &.931 &.055 & .902 & .879  &.910 \\

DCF$_{21}$ (\citeauthor{ji2021calibrated}) & 108.5& 54.3& .024 &.910 & .905 &.941   &  .022&   .918&   .924&   .958
                                      &  .036&   .922&   .912&   .946
                                      &  .039&   .911&   .902&   .940
                                      &  .052&   .899&   .876&   .916 \\

DSA2F$_{21}$ (\citeauthor{sun2021deep})& 36.5 & 172.7 & .021 & .896 & .920  &.962 & .024 & .897 & .918  &.950 & .039 & .901 & .903  &.923 & .036 & .898 & .904  &.933 & - & - & -  &- \\
CMINet$_{21}$ (\citeauthor{cascaded_cmi}) & - & - & .016 & .944 & .940  &.975 & .020 & .931& .932  &.959 & .028 & .940& .929 & .954 & .032 & .925 & .918 &.946 & .040 & .923 & .898  &.934 \\
DSNet$_{21}$ (\citeauthor{wen2021dynamic}) & 172.4 &141.2& .021 & .939  & .928 & .956 & .024 & .925 & .926 & .951 & .034 & .929 & .921 &.946 & .036 & .922 & .914 & .941 & .052 & .899 & .876 & .910\\ 
UTANet$_{21}$ (\citeauthor{zhao2021rgb}) & 48.6& 27.4 & .026 & .921 & .900  &.932 & .020 & .928 & .932  &.964 & .037 & .915 & .902  &.945 & .033 & .921 & .910  &.948 &.048 & .897 & .873  &.925 \\
BIANet$_{21}$ (\citeauthor{zhang2021bilateral}) & 49.6&59.9& .017 & .948 &.942 &.972 & .022 & .926 & .928  &.957 & .034 & .932 & .923  &.945 & .038 & .916 & .908  &.935 & .046 & .908 & .889  &.922  \\
SPNet$_{21}$ (\citeauthor{zhou2021specificity})& 150.3&68.1& .014 & .950 & .945  & .980 &  .021 &  .925 &  .927  &.959 & .028 & .935 & .925  &.954 & .037 & .915 & .907  &.944 & .043 & .916 & .894  &.930 \\ 
VST$_{21}$ (\citeauthor{liu2021visual})& 83.3 & 31.0 &.017&.940&.943&.978&.024&.920&.932&.962&.035&.920&.922&.951&.038&.907&.913&.951&.040&.915&.904&.944\\
RD3D+$_{22}$ (\citeauthor{chen20223})& 28.9 & 43.3 &.017&.946&.950&\highlight{.982}&.022&.921&.933&.964&.033&.928&.928&.955&.037&.905&.914&.946&.046&.900&.892&.928\\
BPGNet$_{22}$ (\citeauthor{yang2022bi})&84.3&138.6&.020&.932&.937&.973&.024&.914&.927&.959&.034&.926&.923&.953&.040&.904&.907&.944&-&-&-&-\\
C2DFNet$_{22}$ (\citeauthor{zhang2022c}) & 47.5&21.0& .020 & .937 & .922 & .948 &.021 & .926 & .928 & .956 & - & -& -& -   & .038 & .911 & .902 & .938 & .053 & .894 & .782 & .911\\ 
MVSalNet$_{22}$ (\citeauthor{zhou2022mvsalnet})   & - &-& .019 & .942& .937& .973  &  .022&   .931&   .930&   .960 &  .036&   .923&   .912&   .944  &  .036&   .921&   .913&   .944 & - & - &- &-\\  
SPSN$_{22}$ (\citeauthor{lee2022spsn})  & 37.0& 100.3& .017 & .942& .937& .973 &  .023&   .917&   .923&   .956  &  .032&   .927&   .918&   .949 &  .035&   .909&   .906&   .941  &  .043&   .910&   .891&   .932 \\   


HiDANet$_{23}$~(\citeauthor{wu2023hidanet})&  130.6&71.5& \highlight{.013} & .952 & .946  &.980 & .021 & .929 & .930  &.961 &  .029 & .939 & .926  &.954 & .035 & .921 & .911  &.946 & .043 & .919 &  .892  & .927\\

\hline
\rowcolor{gray!15}\nMethod{}-T& 5.9 &4.5 & .016 & .947 & .941  &.975 & .021 & .931 & .932  &.960 &  .028 & .937 & .927  &.953 & .033 & .921 &.915  &.945 & .039 & .922 &  .900  & .935\\
\rowcolor{gray!15}\nMethod{}-S&18.5&10.1&.016&.950&.939&.970&.020&.937&.936&.965&.026&.941&.931&.960&.031&.928&.920&.951&.041&.921&.898&.931\\
\rowcolor{gray!15}\nMethod{}-B&29.3&16.7&\highlight{.013}&\highlight{.957}&\highlight{.948}&\highlight{.982}&.019&.933&.936&.965&.025&.941&.933&.960&\highlight{.029}&\highlight{.931}&\highlight{.925}&.951&.035&.932&.908&.943\\
\rowcolor{gray!15}\nMethod{}-L&38.8&26.2&\highlight{.013}&.956&\highlight{.948}&.980&\highlight{.016}&\highlight{.939}&\highlight{.942}&\highlight{.971}&\highlight{.023}&\highlight{.946}&\highlight{.937}&\highlight{.964}&.030&.929&.923&\highlight{.952}&\highlight{.032}&\highlight{.938}&\highlight{.915}&\highlight{.950}\\
\bottomrule
\end{tabular*}}
\vspace{-10pt}
\end{center}
\end{table*}

\subsection{RGB-D Salient Object Detction}\label{sec:exp_sod}

\myPara{Dataset \& implementation details.}
We finetune and test \nMethod{} on five popular RGB-D salient object detection datasets. 
The finetuning dataset consists of 2,195 samples, where 1,485 are from NJU2K-train~\citep{ju2014depth} and the other 700 samples are from NLPR-train~\citep{peng2014rgbd}.
The model is evaluated on five datasets, \ie DES~\citep{cheng2014depth} (135), NLPR-test~\citep{peng2014rgbd} (300), NJU2K-test~\citep{ju2014depth} (500), STERE~\citep{niu2012leveraging} (1,000), and SIP~\citep{fan2020rethinking} (929).
%
For performance evaluation, we adopt four golden metrics of this task, \ie Structure-measure (S)~\citep{fan2017structure}, mean absolute error (M)~\citep{perazzi2012saliency}, max F-measure (F)~\citep{margolin2014evaluate}, and max E-measure (E)~\citep{fan2018enhanced}. 


\begin{table}[t]
  \tablestyle{4pt}{1}
  \begin{minipage}{0.49\linewidth}
  \small
  \caption{RGB-D pretraining. 
    `RGB+RGB' pretraining replaces depth maps with RGB images during pretraining.
    Input channel of the stem layer is modified from 1 to 3.
    The depth map is duplicated three times during finetuning.
  }\label{tab:ablation_ppl}
  \vspace{-10pt}
  \tablestyle{10pt}{1}
  \renewcommand{\arraystretch}{0.8}
  \begin{tabular}{lcc}
    \toprule
    \textbf{pretrain} &\textbf{Finetune}& \textbf{mIoU (\%)} \\
    \midrule\midrule
    RGB+RGB & RGB+D & 53.3 \\
    RGB+D (\textbf{Ours}) & RGB+D & 55.6 \\ \bottomrule
  \end{tabular}\label{tab:rev}
  \end{minipage}
  \small
  \hfill
  \begin{minipage}{0.47\linewidth}
  \vspace{-10pt}
  \small
  \caption{Different inputs of the decoder head for \nMethod{}-B. 
    `$X^{rgb}$+$X^{d}$' means simultaneously uses RGB and depth features. 
    Specifically, both features from the last three stages are used 
    as the input of the decoder head. 
  }\label{tab:decoderin}
  \vspace{-10pt}
  \renewcommand{\arraystretch}{0.8}
  \begin{tabular}{lccc} \toprule
    Decoder input&\textbf{\#Params}&\textbf{FLOPs}&\textbf{mIoU}\textbf{(\%)}\\
    \midrule\midrule
    $X^{rgb}$ (\textbf{Ours})& 29.5 & 41.9 & 55.6 \\
    $X^{rgb}+X^{d}$& 30.8 & 44.8 & 55.5 \\ \bottomrule
  \end{tabular}
  \label{tab:intersup}
  \vspace{-10pt}
  \end{minipage}
\vspace{-10pt}
\end{table}

\begin{figure}[tp]
\centering
\includegraphics[width=\linewidth]{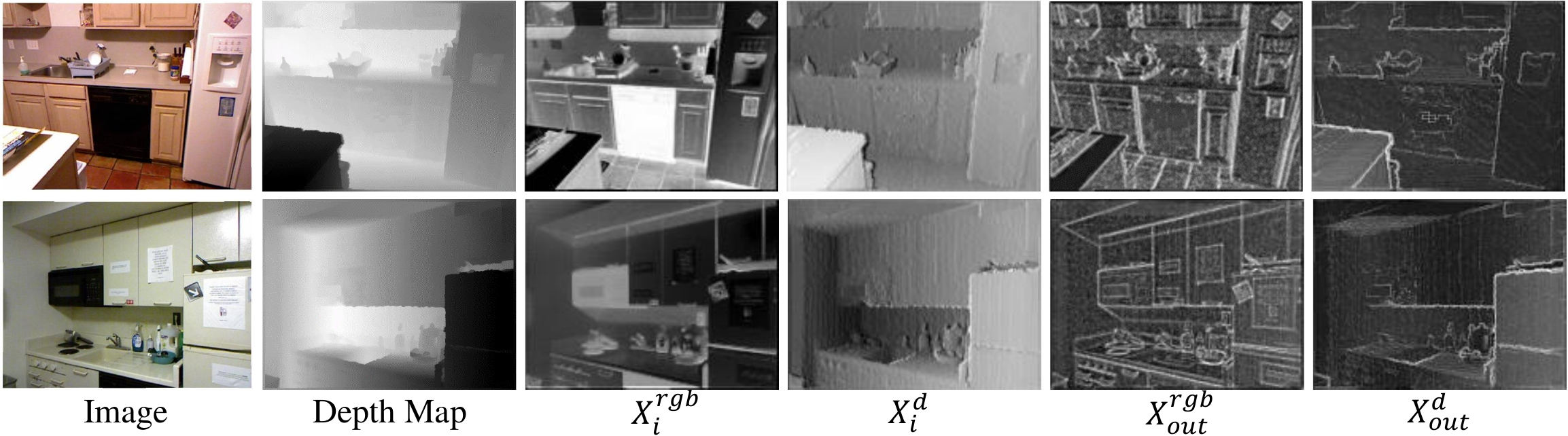}
\vspace{-20pt}
\caption{\small Visualizations of the feature maps around the last RGB-D block of the first stage.
}\label{fig:visualization}
\end{figure}

\myPara{Comparisons with state-of-the-art methods.}
We compare our \nMethod{} with 11 recent RGB-D salient object detection methods on the five popular test datasets.
As shown in \tabref{tab:sod}, our \nMethod{} is able to surpass all competitors with the least computational cost. 
More importantly, our \nMethod{}-T yields comparable performance to the recent \sArt method SPNet~\citep{zhou2021specificity} with less than 10\% computational cost (5.9M, 4.5G vs 150.3M, 68.1G).
The significant improvement further illustrates the strong performance of \nMethod{}.

\subsection{Ablation Study and Analysis}\label{sec:ab}
We perform ablation studies to investigate the effectiveness of each component.
All experiments here are conducted under RGB-D segmentation setting on NYU DepthV2~\citep{silberman2012nyu_dataset}.

\myPara{RGB-D \emph{vs.} RGB pretraining.}
To explain the necessity of the RGB-D pretraining, we attempt to replace the depth maps with RGB images during pretraining, dubbed as RGB pretraining.
To be specific, RGB pretraining modifies the input channel of the depth stem layer from 1 to 3 and duplicates the depth map three times during finetuning.
Note that for the finetuning setting, the modalities of the input data and the model structure are the same.
%
%
As shown in \tabref{tab:ablation_ppl}, our RGB-D pretraining brings 2.3\% improvement for \nMethod{}-B compared to the RGB pretraining in terms of mIoU on NYU DepthV2.
We argue that this is because our RGB-D pretraining avoids the mismatch encoding of the 3D geometry features of depth maps caused by the use of pretrained RGB backbones and enhances the interaction efficiency between the two modalities.
\tabref{tab:dbackbone} and \figref{fig:ab_ppl} in the appendix also demonstrate this.
%
%
These experiment results indicate that the RGB-D representation capacity learned during the RGB-D pretraining is crucial for segmentation accuracy.

%





\myPara{Input features of the decoder.}
Benefiting from the powerful RGB-D pretraining, the features of the RGB branch can efficiently fuse the information of two modalities.
Thus, our decoder only uses the RGB features $X^{rgb}$, 
which contains expressive clues, instead of using both $X^{rgb}$ and $X^{d}$.
As shown in \tabref{tab:decoderin}, using only $X^{rgb}$ can save computational cost without performance drop, while other methods usually need to use both $X^{rgb}$ and $X^{d}$.
This difference also demonstrates that our proposed RGB-D pretraining pipeline and block are more suitable for RGB-D segmentation.


\begin{table}[t]
\vspace{-10pt}
  \tablestyle{6pt}{1}
  \begin{minipage}{0.48\linewidth}
  \caption{Ablation results on the components of the RGB-D block in \nMethod{}-S.
  }\label{tab:ab_block}
  \vspace{-10pt}
  \renewcommand{\arraystretch}{0.7}
  \begin{tabular}{lccc}
    \toprule
    \textbf{Model} & \textbf{\#Params}  & \textbf{FLOPs} & \textbf{mIoU (\%)} \\
    \midrule\midrule
    \nMethod{}-S& 18.7M & 25.6G &53.6  \\
    w/o Base &14.6M&19.6G&52.1\\
    w/o GAA& 16.5M & 21.7G & 52.3  \\
    w/o LEA& 16.3M &23.3G&52.6 \\
    \bottomrule
  \end{tabular}
  \end{minipage}
  \hfill
  \begin{minipage}{0.49\linewidth}
  \caption{Ablation on GAA in the \nMethod{}-S.   
    `$k \times k$' means the adaptive pooling size of $Q$. 
  }\label{tab:fixed_size}
  \vspace{-10pt}
  \tablestyle{4pt}{1}
  \renewcommand{\arraystretch}{0.7}
  \begin{tabular}{cccc} \toprule
    \textbf{Kernel Size}&\textbf{\#Params}&\textbf{FLOPs}&\textbf{mIoU (\%)} \\ \midrule\midrule
    3$\times$3&18.7M&22.0G&52.7\\
    5$\times$5&18.7M&23.3G&53.1\\
    7$\times$7&18.7M&25.6G&53.6\\
    9$\times$9&18.7M&28.8G&53.6\\ \bottomrule
  \end{tabular}
  \end{minipage}
  \begin{minipage}{0.48\linewidth}
  \vspace{2pt}
  \caption{Different fusion methods for $Q$, $K$, and $V$.
  We can see that there is no need to fuse RGB and depth features for $K$ and $V$.
  }\label{tab:gaa}
  \vspace{-10pt}
  \tablestyle{3.4pt}{1}
  \renewcommand{\arraystretch}{0.8}
  \begin{tabular}{cccc} \toprule
    \textbf{Fusion variable} & \textbf{\#Params}  & \textbf{FLOPs} & \textbf{mIoU (\%)} \\ \midrule\midrule
    None&18.5M&25.4G&53.2\\
    only $Q$ (ours)& 18.7M  & 25.6G  & 53.6  \\
    $Q$, $K$, $V$& 19.3M & 28.5G &  53.6 \\ \bottomrule
  \end{tabular}
  \end{minipage}
  \hfill
  \begin{minipage}{0.49\linewidth}
  \vspace{2pt}
  \caption{Different fusion manners in LEA module. 
    Fusion manner refers to the operation to fuse the RGB and depth 
    information in LEA (\eqnref{eq:A}).  
  }\label{tab:lse}
  \vspace{-10pt}
  \tablestyle{5pt}{1}
  \renewcommand{\arraystretch}{0.8}
  \begin{tabular}{lccc} \toprule
    Fusion manner& \textbf{\#Params}  & \textbf{FLOPs} & \textbf{mIoU}\textbf{(\%)} \\ \midrule\midrule
    Addition& 18.6M & 25.1G & 53.1 \\
    Concatenation& 19.0M & 26.4G & 53.3 \\
    Hadamard&18.7M&25.6G&53.6\\ \bottomrule
  \end{tabular}
  \end{minipage}
  \vspace{-5pt}
\end{table}

\myPara{Components in our RGB-D block.}
Our RGB-D block is composed of a base module, GAA module, and LEA module.
We take out these three components from \nMethod{} respectively, where the results
are shown in \tabref{tab:ab_block}.
It is clear that all of them are essential for our \nMethod{}.
Moreover, we visualize the features around the RGB-D block in \figref{fig:visualization}.
It can be seen the output features can capture more comprehensive details.  
As shown in \tabref{tab:gaa}, in the GAA module, we find that only fusing the depth features into $Q$ is adequate and further fusing depth features to $K$ and $V$ brings negligible improvement but extra computational burdens.
Moreover, we find that the performance of \nMethod{} initially rises as the fixed pooling size of the GAA increases, and it achieves the best when the fixed pooling size is set to 7×7 in \tabref{tab:fixed_size}. 
%
%
%
We also use two other fusion manners to replace that in LEA, \ie concatenation, addition.
As shown in \tabref{tab:lse}, using the depth features that processed by a large kernel depth-wise convolution as attention weights to reweigh the RGB features via a simple Hadamard product achieves the best performance.
%

\begin{wrapfigure}{r}{0.45\textwidth}
\vspace{-9.5pt}
\centering
\includegraphics[width=0.85\linewidth]{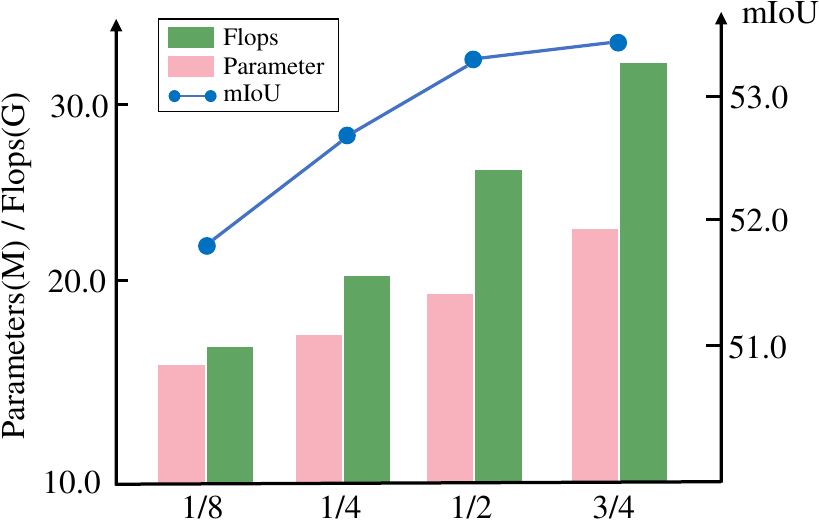}
\vspace{-10pt}
\caption{\footnotesize Performance on different channel ratios $C^{d}/C^{rgb}$ based on \nMethod{}-S. $C^{rgb}$ is fixed and we adjust $C^{d}$ to get different ratios.
}\label{fig:ratio}
\vspace{-15pt}
\end{wrapfigure}

\myPara{Channel ratio between RGB and depth.}
RGB images contain information pertaining to object color, texture, shape, and its surroundings. 
In contrast, depth images typically convey the distance information from each pixel to the camera.
Here, we investigate the channel ratio that is used to encode the depth information.
In \figref{fig:ratio}, we present the performance of \nMethod{}-S with different channel ratios, \ie $C^{d}/C^{rgb}$.
We can see that when the channel ratio excceds `1/2', the improvement is trivial while the computational burden is significantly increased.
Therefore, we set the ratio to 1/2 by default.

\begin{table}[tp]
  \tablestyle{4pt}{1}
  \begin{minipage}{0.49\linewidth}
  \small
  \caption{Comparison under the RGB-only pretraining. `NYU' and `SUN' means the performance on the NYU DepthV2 and SUNRGBD.
  }\label{tab:ablation_pre}
  \vspace{-10pt}
  \renewcommand{\arraystretch}{0.8}
  \begin{tabular}{lcccc} \toprule
    Model&\textbf{Params}&\textbf{FLOPs}&NYU&SUN\\
    \midrule\midrule
    CMX (MiT-B2)&66.6M & 67.6G & 54.4 & 49.7 \\
    \rowcolor{gray!15}\nMethod{}-B& 29.5M & 41.9G & 53.3 & 49.5  \\
    \rowcolor{gray!15}\nMethod{}-L& 39.0M & 65.7G & 55.4 &50.6 \\ \bottomrule
  \end{tabular}\label{tab:rev1}
  \end{minipage}
  \small
  \hfill
  \begin{minipage}{0.47\linewidth}
  \vspace{-30pt}
  \small
  \caption{Comparison under the RGB-D pretraining. `NYU' and `SUN' means the performance on the NYU DepthV2 and SUNRGBD.
  }\label{tab:onlyRGB}
  \vspace{-10pt}
  \renewcommand{\arraystretch}{0.8}
  \begin{tabular}{lcccc} \toprule
    Model&\textbf{Params}&\textbf{FLOPs}&NYU&SUN\\
    \midrule\midrule
    CMX (MiT-B2)&66.6M & 67.6G & 55.8 & 51.1 \\
    \rowcolor{gray!15}\nMethod{}-B&29.5M & 41.9G &55.6 & 51.2 \\
    \rowcolor{gray!15}\nMethod{}-L& 39.0M & 65.7G & 57.2 & 52.5 \\ \bottomrule
  \end{tabular}
  \label{tab:preRGBD}
  \vspace{-30pt}
  \end{minipage}
\vspace{-20pt}
\end{table}

\myPara{Apply the RGB-D pretraining manner to CMX.}
To verify the effect of the RGB-D pretraining on other methods and make the comparison more fair, we pretrain the CMX (MiT-B2) on RGB-D data of ImageNet and it obtains about 1.4\% mIoU improvement, as shown in \tabref{tab:preRGBD}.
Under the RGB-D pretraining, DFormer-L still outperforms  CMX (MiT-B2) by a large margin, which should be attributed to that the pretrained fusion weight within DFormer can achieve better and efficient fusion between RGB-D data.
Besides, we provide the RGB pretrained DFormers to provide more insights in \tabref{tab:rev1}.
The similar situation appears under the RGB-only pretraining.

 \begin{wraptable}{r}{0.5\textwidth}
 \small
 \vspace{-14pt}
 \caption{Results on the RGB-T semantic segmentation benchmark MFNet~\citep{ha2017mfnet} and RGB-L semantic segmentation benchmark KITTI-360~\citep{liao2021kitti360}. `(RGB)' and `(RGBD)' mean the RGB-only and RGB-D pretraining, respectively. }
 \vspace{-5pt}
	\resizebox{0.99\linewidth}{!}{\begin{tabular}{lcccc} \toprule
    Model& \textbf{Params}  & \textbf{FLOPs} &MFNet & KITTI\\ \midrule\midrule
     CMX-B2 & 66.6M & 67.6G & 58.2 & 64.3 \\
 CMX-B4 & 139.9M & 134.3G & 59.7 & 65.5* \\
 CMNeXt-B2 & 65.1M & 65.5G & 58.4* &65.3\\
 CMNeXt-B4 &135.6M & 132.6G & 59.9 & 65.6* \\
 \rowcolor{gray!15}Ours-L (RGB) & 39.0M & 65.7G & 59.5 & 65.2 \\
\rowcolor{gray!15} Ours-L (RGBD) & 39.0M & 65.7G & 60.3 & 66.1 \\
 \bottomrule
  \end{tabular}\label{tab:generalization}}
   \vspace{-10pt}
\end{wraptable}

\myPara{Dicussion on the generalization to other modalities.}
Through RGB-D pretraining, the DFormer is endowed with the capacity to interact the RGB and depth during pretraining.
To verify whether the interaction is still work when replace the depth with another modality, we apply our DFormer to some benchmarks with other modalities, \ie RGB-T on MFNet~\citep{ha2017mfnet} and RGB-L on KITTI-360~\citep{liao2021kitti360}.
As shown in the \tabref{tab:generalization} (comparison to more methods are in the \tabref{tab:MFNet} and \tabref{tab:KITTI}), RGB-D pretraining still improves the performance on the RGB and other modalities, nevertheless, the improvement is limited compared to that on RGB-D scenes.
To address this issue, a foreseeable solution is to further scale the pretraining of DFormer to other modalities.
There are two ways to solve the missing of large-scale modal dataset worth trying, \ie synthesizing the pseudo modal data, and seperately pretraining on single modality dataset.
As far as the former, there are some generation methods to generate other pseudo modal data.
For example, Pseudo-lidar~\citep{wang2019pseudo} propose a method to generate the pesudo lidar data from the depth map, and N-ImageNet~\citep{kim2021n} obtain the event data on the ImageNet.
Besides, collecting data and training the modal generator for more modalities, is also worth exploring.
For the latter one, we can separately pretrain the model for processing the supplementary modality and then combine it with the RGB model. 
We will attempt these methods to bring more significant improvements for DFormer on more multimodal scenes.

\section{Related Work}
\vspace{-5pt}

\myPara{RGB-D Scene Parsing}
In recent years, with the rise of deep learning technologies, \emph{e.g.}, CNNs~\citep{he2016resnet}, and Transformers~\citep{vaswani2017attention,li2023sere}, significant progress has been made in scene parsing~\citep{xie2021segformer,yin2022camoformer,chen2023yolo,zhang2023referring}, one of the core pursuits of computer vision. 
However, most methods still struggle to cope with some challenging scenes in the real world~\citep{li2023enhancing,sun2020exploring}, as they only focus on RGB images that provide them with distinct colors and textures but not 3D geometric information. 
To overcome these challenges, researchers combine images with depth maps for a comprehensive understanding of scenes.

Semantic segmentation and salient object detection are two active areas in RGB-D scene parsing. 
Particularly, the former aims to produce per-pixel category prediction across a given scene, and the latter attempts to capture the most attention-grabbing objects.
To achieve the interaction and alignment between RGB-D modalities, the dominant methods investigate a lot of effort in building fusion modules to bridge the RGB and depth features extracted by two parallel pretrained backbones.
For example, methods like CMX~\citep{zhang2022cmx}, TokenFusion~\citep{wang2022multimodal}, and HiDANet~\citep{wu2023hidanet} dynamically fuse the RGB-D representations from RGB and depth encoders and aggregate them in the decoder.
%
%
Indeed, the evolution of fusion manners has dramatically pushed the performance boundary in these applications of RGB-D scene parsing.
Nevertheless, the three common issues, as discussed in Sec.~\ref{sec:intro}, are still left unresolved.
%
Another line of work focuses on the design of operators~\citep{wang2018depth,wu2020depth,cao2021shapeconv,chen2021spatial_guided} to extract complementary information from RGB-D modalities. 
For instance, methods like ShapeConv~\citep{cao2021shapeconv}, SGNet~\citep{chen2021spatial_guided}, and Z-ACN~\citep{wu2020depth} propose depth-aware convolutions, which enable efficient RGB features and 3D spatial information integration to largely enhance the capability of perceiving geometry.
Although these methods are efficient, the improvements brought by them are usually limited due to the insufficient extraction and utilization of the 3D geometry information involved in the depth modal.

\myPara{Multi-modal Learning}
The great success of the pretrain-and-finetune paradigm in natural language processing and computer vision has been expanded to the multi-modal domain, and the learned transferable representations have exhibited remarkable performance on a wide variety of downstream tasks.
Existing multi-modal learning methods cover a large number of modalities, \emph{e.g.}, image and text~\citep{castrejon2016learning,chen2020uniter,radford2021learning,zhang2021rstnet,wu2022difnet}, text and video~\citep{akbari2021vatt}, text and 3D mesh~\citep{zhang2023temo}, image, depth, and video~\citep{girdhar2022omnivore}.
These methods can be categorized into two groups, \emph{i.e.}, multi- and joint-encoder ones.
Specifically, the multi-encoder methods exploit multiple encoders to independently project the inputs in different modalities into a common space and minimize the distance or perform representation fusion between them.
For example, methods like CLIP~\citep{radford2021learning} and VATT~\citep{akbari2021vatt} employ several individual encoders to embed the representations in different modalities and align them via a contrastive learning strategy.
%
In contrast, the joint-encoder methods simultaneously input the different modalities and use a multi-modal encoder based on the attention mechanism to model joint representations.
For instance, MultiMAE~\citep{bachmann2022multimae} adopts a unified transformer to encode the tokens with a fixed dimension that are linearly projected from a small subset of randomly sampled multi-modal patches and multiple task-specific decoders to reconstruct their corresponding masked patches by the attention mechanism separately.

In this paper, we propose \nMethod{}, a novel framework that achieves RGB-D representation learning in a pretraining manner.
To the best of our knowledge, this is the first attempt to encourage the semantic cues from RGB and depth modalities to align together by the explicit supervision signals of classification, yielding transferable representations for RGB-D downstream tasks.

\section{Conclusions}

In this paper, we propose a novel RGB-D pretraining framework to learn transferable representations for RGB-D downstream tasks. 
%
%
Thanks to the tailored RGB-D block, our method is able to achieve better interactions between the RGB and depth modalities during pretraining.
Our experiments suggest that \nMethod{} can achieve new state-of-the-art performance in RGB-D downstream tasks, \eg semantic segmentation and salient object detection, with far less computational cost compared to existing methods.

\section*{Acknowledgments}
This research was supported by National Key Research and Development Program of China (No. 2021YFB3100800), NSFC (NO. 62225604, No. 62276145, and No. 62376283), the Fundamental Research Funds for the Central Universities (Nankai University, 070-63223049), CAST through Young Elite Scientist Sponsorship
Program (No. YESS20210377). Computations were supported by the Supercomputing Center of Nankai University (NKSC).

\section*{Reproducibility Statement}

Ensuring the reproducibility of our research is important to us. 
In this reproducibility statement, we outline the measures taken to facilitate the replication of our work and provide references to the relevant sections in the main paper, appendix, and the code.

\myPara{Source code.}
We have made our source code anonymously available in \href{https://github.com/VCIP-RGBD/DFormer}{Link}, allowing researchers to access and utilize our code for reproducing our experiments and results.
Detailed installation instructions are in `README.md.'
The source code and model weights will be made public available.

\myPara{Experimental setup.}
In the main paper, we provided basic parameter settings and implementation settings in \secref{sec:RGBDPretrain} (pretraining), \secref{sec:exp_seg} (RGB-D semantic segmentation), and \secref{sec:exp_sod} (RGB-D salient object detection).
In the \tabref{tab:conf} of the appendix, we provide the detailed configuration for different variants of our \nMethod{}.
We provide the training details in \tabref{tab:train_detail} and \tabref{tab:fine_set}.
Moreover, the experimental setups can be seen in the source code in the supplementary materials.


By providing these resources and references, we aim to enhance the reproducibility of our work and enable fellow researchers to verify and build upon our findings. We welcome any inquiries or requests for further clarification on our methods to ensure the transparency and reliability of our research.

%
%


\bibliography{iclr2024_conference}
\bibliographystyle{iclr2024_conference}

\newpage

\section*{Appendix}
\appendix

In \secref{sec:analysis}, we first provide further analysis of our \nMethod{}: 1) the efficiency in encoding the depth maps; 2) the effect of depth maps in different quality during pretraining.
Then we present more detailed descriptions of \nMethod{} in \secref{sec:structure} and the experimental settings in \secref{sec:setting}.
Finally, we provide more visualization results of \nMethod{} in \secref{sec:vis} and future exploring directions in \secref{sec:future}.

\section{More analysis of \nMethod{}}\label{sec:analysis}

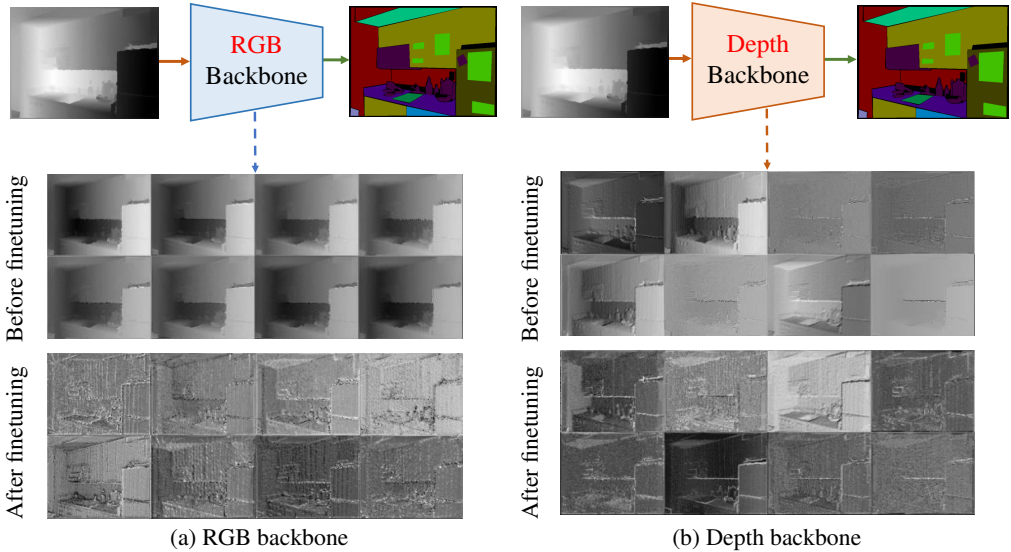
\begin{figure}[ht]
\centering
\vspace{5pt}
\includegraphics[width=0.95\linewidth]{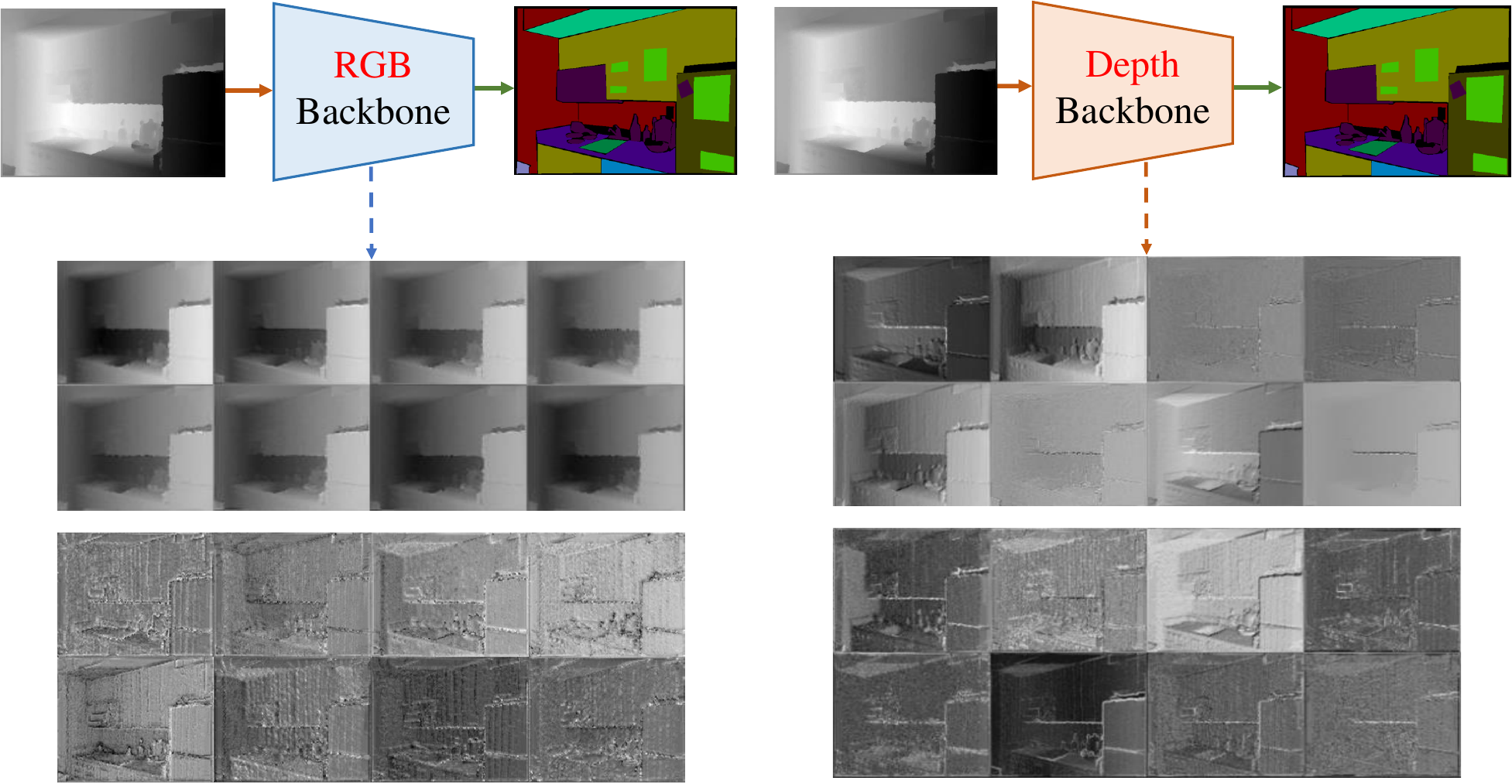}
\put (-379, 0){\small \rotatebox{90}{After finetuning}}
\put (-379, 67){\small \rotatebox{90}{Before finetuning}}
\put (-184, 0){\small \rotatebox{90}{After finetuning}}
\put (-184, 67){\small \rotatebox{90}{Before finetuning}}
\put (-317, -10){\small (a) RGB backbone}
\put (-127, -10){\small (b) Depth backbone}
\vspace{-5pt}
\caption{Encoding depth maps with backbones pretrained on different types of training data. (a) Pretraining with RGB data. (b) Pretraining with depth maps. During finetuning, we only take the depth maps as input to see which backbone works better.
We visualize some features from the two backbones as shown in the bottom part.
Obviously, the backbone pretrained on depth maps can generate more expressive feature maps.
%
}\label{fig:ab_ppl}
\vspace{-1pt}
\end{figure}

\begin{table}[ht]
\captionsetup{font={small}}
\vspace{10pt}
\caption{Performance of the RGB pretrained and depth pretrained backbone processing depth maps for segmentation. The two backbones adopt the same pretraining setting and architecture but are pretrained on the ImageNet images and their depth maps respectively.}\label{tab:dbackbone}
\vspace{-5pt}
\renewcommand\arraystretch{1.0}
\setlength{\tabcolsep}{7pt}
\centering
\small
\begin{tabular}{lccc}
    \toprule
    Backbone& \textbf{\#Params}  & \textbf{FLOPs} & \textbf{mIoU}\textbf{(\%)} \\
    \midrule\midrule
    RGB&11.2M& 13.7G&27.6\\
    Depth&11.2M&13.7G &42.8\\
    \bottomrule
\end{tabular}
\end{table}

\myPara{Why involving depth information in pretraining?}
The existing \sArt RGB-D methods~\citep{zhang2022cmx,wang2022multimodal} tend to use models pretrained on RGB images to encode 3D geometry information for depth maps.
We argue that the huge representation distribution shift caused by using RGB backbone to encode the depth maps may influence the extraction of the 3D geometry. 
%
%
To demonstrate this, we respectively use RGB and depth data to pretrain the RGB and depth backbone and then take only the depth maps as input for segmentation, as shown in \figref{fig:ab_ppl}.
%
From \tabref{tab:dbackbone}, we can see under the same network architecture, the model pretrained on depth images performs much better than the one pretrained on RGB images, \ie yielding an improvement of 15\% mIoU.
%
%
To delve into the reasons, we visualize the feature maps as shown in the bottom part of \figref{fig:ab_ppl}.
Before fing, the backbone pretrained on RGB data is not able to extract expressive features from the depth maps.
After fing, the model using RGB backbones still struggles to extract diverse features from the depth maps.
On the contrary, the features from the backbone pretrained on depth data are better.
These experiments indicate that there exists significant difference RGB and depth maps and it is difficult to process depth data with RGB pretrained weights.
%
%
%
This also motivates us to involve depth data during ImageNet pretraining.


\begin{figure}[ht]
\centering
\includegraphics[width=0.96\linewidth]{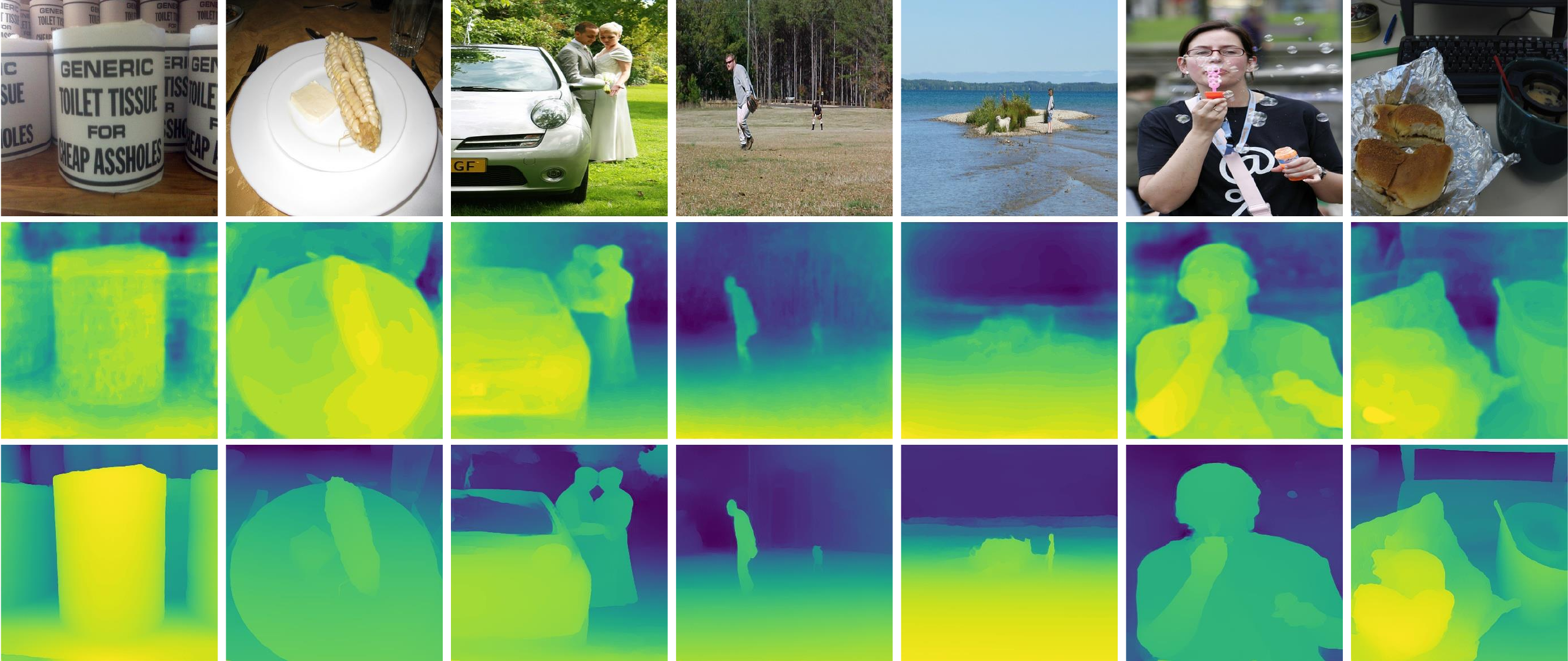}
\put (-393, 125){\small \rotatebox{90}{Image}}
\put (-393, 68){\small \rotatebox{90}{Adabins}}
\put (-393, 10){\small \rotatebox{90}{Omnidata}}
\caption{\small Comparison of esitimated depth maps that generated by Adabins~\citep{bhat2021adabins} and more advanced Omnidata~\citep{eftekhar2021omnidata}. We visualize the depth maps in color for better comparison.
}\label{fig:depth_compare}
\end{figure}

\myPara{Impact of the quality of depth maps.}
In the main paper, we use a depth estimation method Adabins~\citep{bhat2021adabins} to predict depth maps for the ImageNet-1K~\citep{russakovsky2015imagenet} dataset.
Both the ImageNet images and the generated depth maps are used to pretrain our \nMethod{}.
Here, we explore the effect of depth maps in different quality on the performance of \nMethod{}.
To this end, we also choose a more advanced depth estimation method, \ie Omnidata~\citep{eftekhar2021omnidata}, to generate the depth maps.
In \figref{fig:depth_compare}, we visualize the generated depth maps that are generated by these two methods.
We use the Omnidata-predicted depth maps to replace the Adabins-predicted ones during RGB-D pretraining.
Then, we use this to pretrain the DFormer-Base and it achieves nearly the same performance, \ie 55.6 mIoU on NYUDepth v2.
%
%
This indicates the quality of depth maps has little effect on the performance of \nMethod{}.
%
%

\begin{figure}[ht]
\centering
\includegraphics[width=0.99\linewidth]{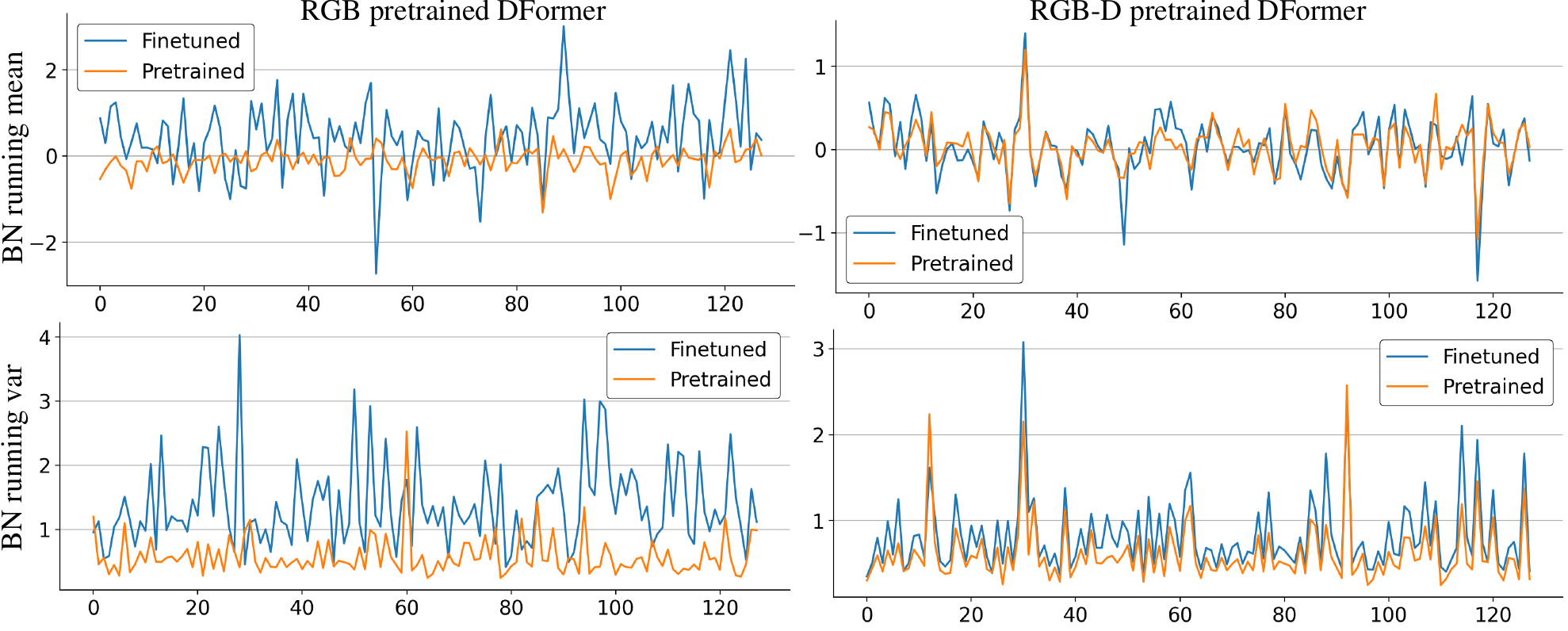}
\vspace{-10pt}
\caption{\small The statistics of distribution shift on the finetuning of DFormer-S that uses different pretraining manners. The BN layer at the first block of stage 2 is chosen for this visualization.}
\label{fig:zxt_dis}
\end{figure}

\begin{figure}[tp]
\centering
\includegraphics[width=0.99\linewidth]{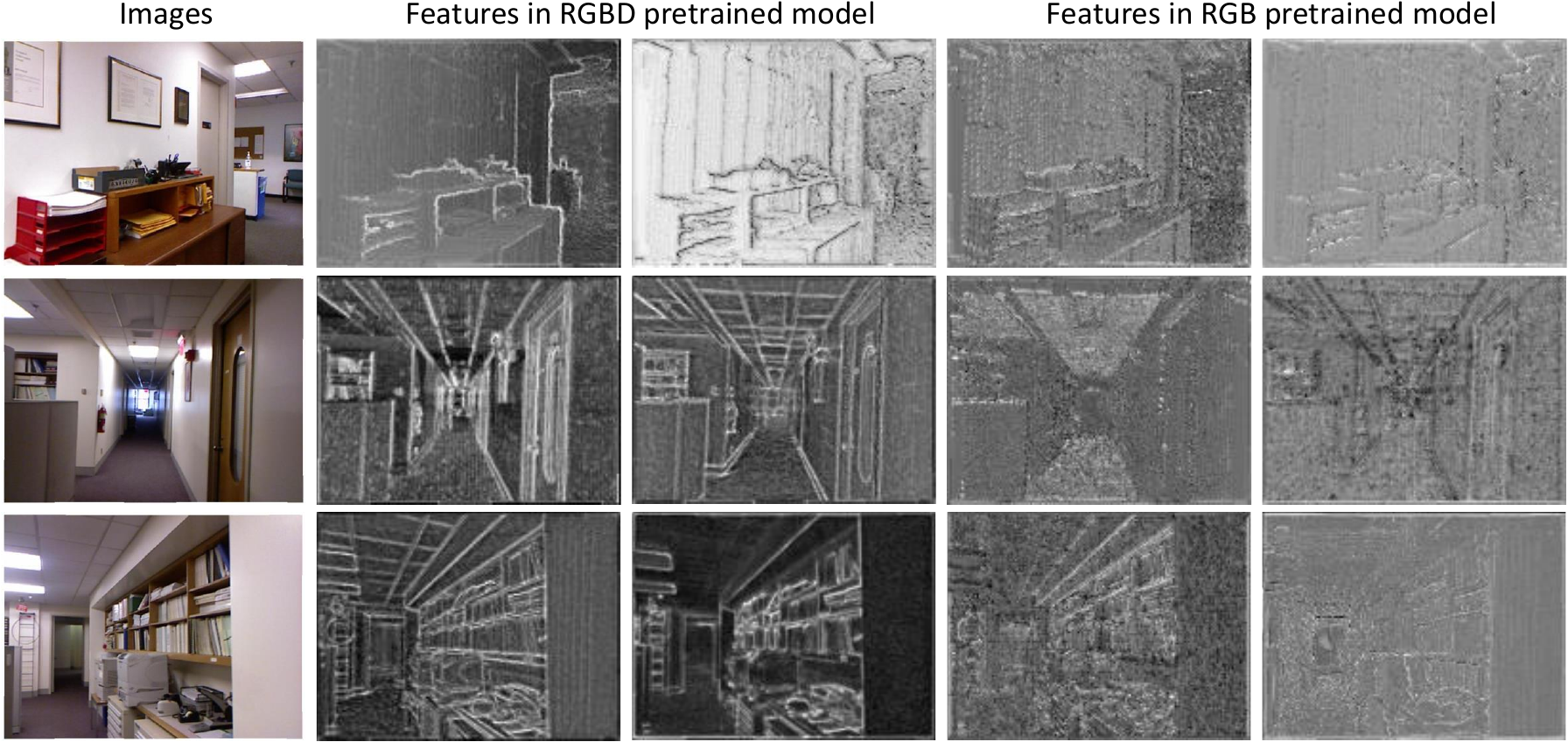}
\vspace{-10pt}
\caption{\small Visualization of the features within the finetuned models that load RGB-pretrained and RGBD-pretrained weights. The features are randomly picked from the first stage output in the \nMethod{}-S.
}\label{fig:feat_reb}
\end{figure}

\myPara{Observations towards the distribution shift.}
Interacting the RGB and depth features within the RGB-pretrained encoders would bring in drastic changing of the feature distribution, which makes the previous statistic of batch normalization incompatible with the input features.
Folowing the~\citep{chen2021empirical}, we visualize the statistics of the BN layers during the finetuning to reflect the distribution shift.
Specifically, we visualize the statistics of batch normalization for a random layer in the \nMethod{} in \figref{fig:zxt_dis} of the new revision to observe the statistic of the fused features to the RGB backbone.
For the RGBD-pretrained DFormer, the running mean and variance of the BN layer only have slight changes after finetuning, illustrating the learned RGBD representation is transferable for the RGBD segmentation tasks.
In contract, for the RGB-pretrained one, the statistics of the BN layer are indeed drastically changed after finetuning, which indicates the encoding is mismatched.
The situation forces RGB-pretrained weights to adapt the input features fused by two modalities.
We also visualize some features of the DFormer that uses the RGB and RGBD pretraining, as shown in \figref{fig:feat_reb}.

\section{Details of \nMethod{}} \label{sec:structure}

\begin{figure}[ht]
\centering
\vspace{5pt}
\includegraphics[width=0.7\linewidth]{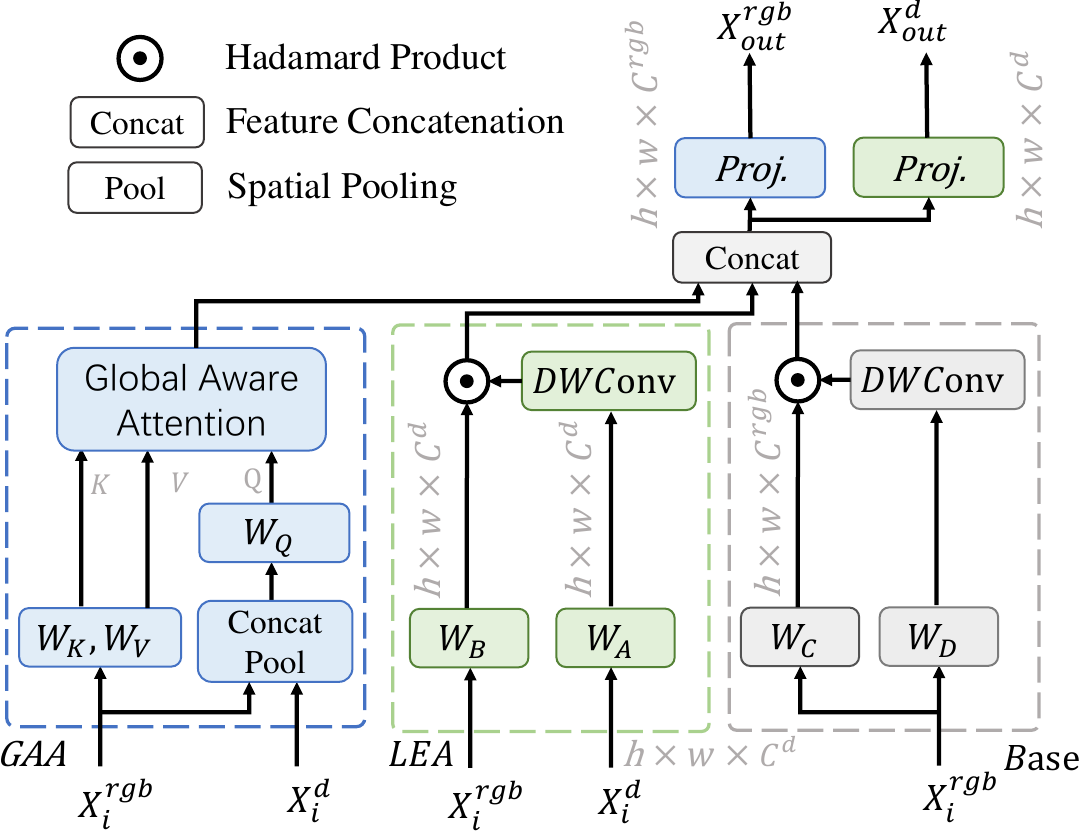}
\vspace{-5pt}
\caption{Detailed structure of our RGB-D block in \nMethod{}. 
}\label{fig:detaileds}
\vspace{10pt}
\end{figure}

\myPara{Structure.} In the main paper, we only present the structure map of the interaction modules due to the limited space, and the base modules is omitted.
The detailed structure of our RGB-D block is presented in \figref{fig:detaileds}.
The GAA, LEA and base modules jointly construct the block, and each of them is essential and contribute to the performance improvement.
GAA and LEA aims to conduct interactions between different modalities globally and locally, while the base module is only responsible for encoding RGB features.
As the first stage focuses on encoding low-level feature, we do not use GAA in the first stage for the sake of efficiency.
Moreover, the MLP layers for the RGB and depth features are indivisual.

\begin{wrapfigure}{r}{0.49\textwidth}
\vspace{-11.5pt}
\centering
\setlength{\abovecaptionskip}{2pt}
\includegraphics[width=0.99\linewidth]{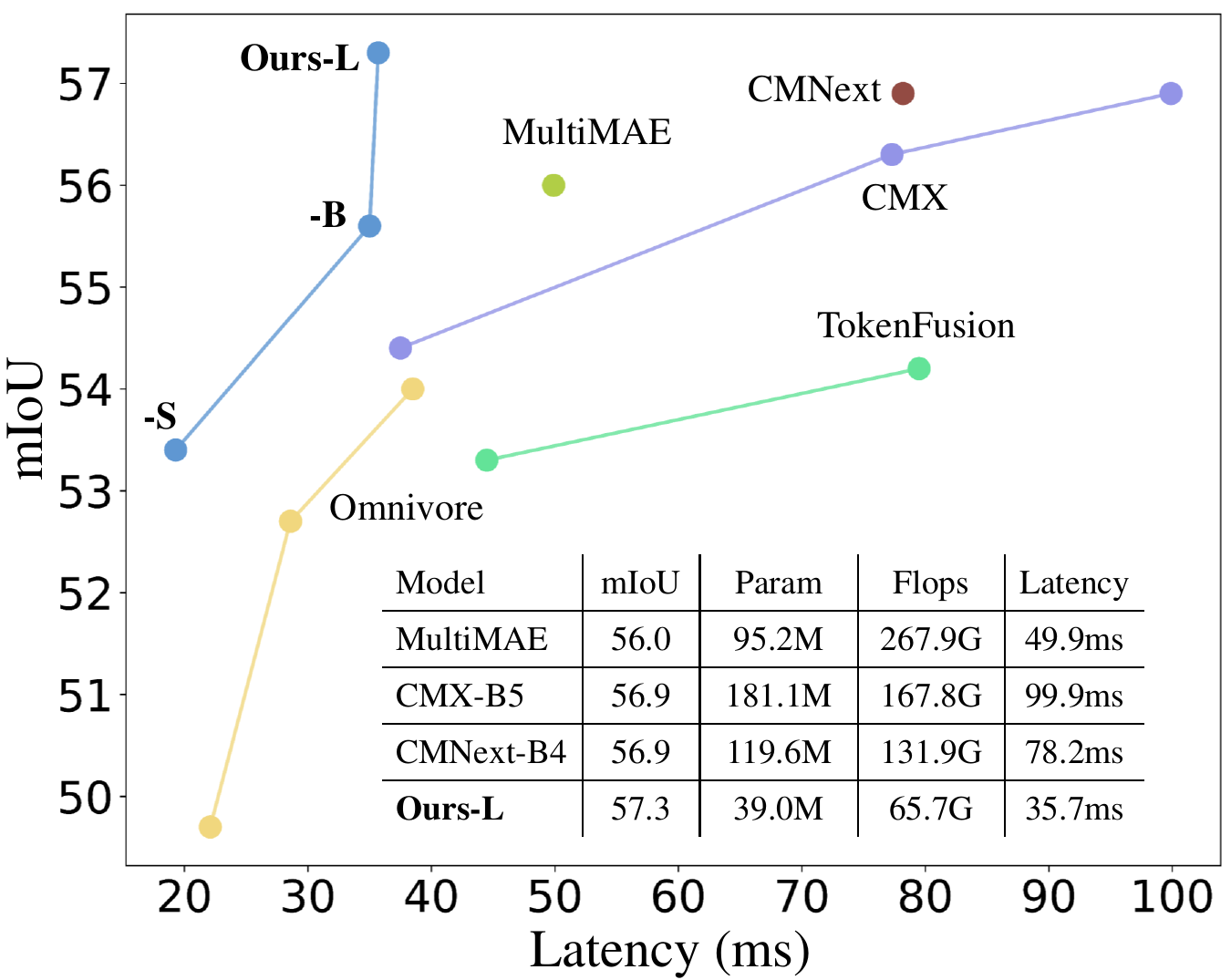}
\caption{\footnotesize Performance vs. Latency when processing 480$\times$640 images. 
}\label{fig:latency}
\vspace{-15pt}
\end{wrapfigure}

\myPara{Inference time analysis.}
Real-time inference of an RGB-D model plays a key role in a wide spectrum of downstream applications, as stated by~\citep{chen2020sa_gate}.
To this end, we conduct experiments to explore the real-time potential of our \nMethod{} and other methods.
To ensure fairness, all comparisons are performed on the same device, \ie a single 3090 RTX GPU, and the same image resolution, \ie $480\times 640$.
As illustrated in \figref{fig:latency}, our \nMethod{}-L achieves 57.2\% mIoU with 35.7 ms latency, while the latency of current \sArt CMNext is 78.2 ms. 
Remarkably, our \nMethod{}-S can process an image in 20ms and achieve about 50 frames per second (FPS) with competitive performance on NYU Depthv2, \ie 53.6\% mIoU.

\begin{table}[ht]
\captionsetup{font={small}}
\vspace{10pt}
\caption{Ablation towards the base modules within the building block of \nMethod{}-S.}\label{tab:base_ab}
\vspace{-5pt}
\renewcommand\arraystretch{1.0}
\setlength{\tabcolsep}{7pt}
\centering
\small
\begin{tabular}{lcccc} \toprule
DWConv Setting&Attention Operation&Param&Flops&NYUDepthV2\\ 
    \toprule
    DWConv $7\times7$&Hadamard Product&18.7M&25.6G&51.9\\ 
DWConv $5\times5$&Hadamard Product&18.7M&23.9G&51.6\\
DWConv $9\times9$&Hadamard Product&18.7M&27.1G&51.9\\ 
DWConv $7\times7$&Addition&18.7M&25.0G&51.3\\
DWConv $7\times7$&Concatanation&19.3M&26.9G&51.7\\
    \bottomrule
\end{tabular}
\end{table}

\myPara{More detailed ablation towards the block.}
In the main paper, we have provided the ablation experiments about the components of our RGB-D block, the pooling size of our GAA, as well as the fusion manners in GAA and LEA.
Here we provide more results for the modules that encode RGB features, as shown in \tabref{tab:base_ab}.
Due to limited time and computation resources, we use a short pretraining duration of 100 epochs on the DFormer-S.
Note that the results in the below table only ablates the stucture within the modules (gray part in \figref{fig:detaileds}) that only process the RGB features.

\newcommand{\MCols}[2]{\multicolumn{#1}{c}{#2}}
\begin{table}[t]
  \tablestyle{3.7pt}{1.25}
  \caption{Detailed configurations of the proposed \nMethod{}.
    `$C$' = (`$C_{rgb}$',`$C_{d}$'), 
    which respectively represent the channel number of the RGB part and the depth part in different stages.  
    `$N_i$' is the number of building blocks in $i$-th stage.
    `Expansion' is the expand ratio for the number of channels in MLPs. 
    `Decoder dimension' denotes the channel dimension in the decoder. 
  }\label{tab:conf}
  \vspace{-8pt}
  \small
  \begin{tabular}{cccccccc} \toprule
    Stage & Output size  & Expansion & \nMethod{}-T & \nMethod{}-S & \nMethod{}-B & \nMethod{}-L \\ \midrule
    Stem & $\frac{H}{4}\times \frac{W}{4} $ & - &\makecell{$C=(16,8)$}&\makecell{$(32,16)$}&\makecell{$(32,16)$}&\makecell{$(48,24)$} \\
    1 & $\frac{H}{4}\times \frac{W}{4} $ & 8  & 
    \makecell{$C=(32,16)$, $N_{1}=3$} & \makecell{$(64,32)$, $2$} & \makecell{$(64,32)$, $3$} & $(96,48), 3$ \\ 
    2 & $\frac{H}{8}\times \frac{W}{8} $ & 8  & 
    \makecell{$C=(64,32)$, $N_{2}=3$} & \makecell{$(128,64)$, $2$} & \makecell{$(128,64)$, $3$} &  \makecell{$(192,96)$, $3$} \\ 
    3 & $\frac{H}{16}\times \frac{W}{16} $ & 4 & 
    \makecell{$C=(128,64)$, $N_{3}=5$} & \makecell{$(256,128), 4$} & \makecell{$(256,128)$, $12$} &  \makecell{$(288,144)$, $12$} \\ 
    4 & $\frac{H}{32}\times \frac{W}{32} $ & 4 & 
    \makecell{$C=(256,128)$, $N_{4}=2$} & \makecell{$(512,256)$, $2$} & \makecell{$(512,256)$, $2$} &  \makecell{$(576,288)$, $3$} \\ \midrule
    \MCols{3}{Decoder dimension} & 512 & 512 & 512 & 512 \\ 
    \MCols{3}{{Parameters} (M)} & 6.0 & 18.7  & 29.5  & 39.0 \\ 
    \bottomrule
  \end{tabular}
  \vspace{-10pt}
\end{table}

\subsection{Comparison on the MFNet and KITTI-360}

For a more comprehensive comparison, we compare our \nMethod{} and more other \sArt methods on the MFNet~\citep{ha2017mfnet} and KITTI-360~\citep{liao2021kitti360} in \tabref{tab:KITTI} and \tabref{tab:MFNet}, as a supplement to the main paper.

    

\begin{table}[tp]
  \tablestyle{4pt}{1}
  \begin{minipage}{\linewidth}
  \centering
  \small
  \caption{MFNet (RGB-T)~\citep{ha2017mfnet}.}
    \vspace{-5pt}
    \vskip -1ex
    \label{tab:table_SUN}
    \setlength{\tabcolsep}{2pt} 
    \renewcommand{\arraystretch}{0.9}
    	\begin{tabular}{l|cc}
        \toprule
        \textbf{Method} &\textbf{Backbone} & \textbf{mIoU (\%) }  \\
        \midrule\midrule
        ACNet~(\citeauthor{hu2019acnet}) & ResNet-50   & 46.3 \\
        FuseSeg~(\citeauthor{sun2020fuseseg}) & DenseNet-161 & 54.5  \\
        ABMDRNet~(\citeauthor{zhang2021abmdrnet})  & ResNet-18 & 54.8 \\
        LASNet~(\citeauthor{li2022rgb}) & ResNet-152 & 54.9 \\
        FEANet~(\citeauthor{deng2021feanet}) & ResNet-152   & 55.3 \\
        MFTNet~(\citeauthor{zhou2022multispectral}) & ResNet-152 & 57.3\\
        GMNet~(\citeauthor{zhou2021gmnet})  & ResNet-50 & 57.3 \\
        DooDLeNet~(\citeauthor{frigo2022doodlenet}) & ResNet-101 &57.3 \\
        CMX~(\citeauthor{zhang2022cmx}) & MiT-B2 & 58.2\\
        CMX~(\citeauthor{zhang2022cmx}) & MiT-B4 & 59.7\\
        CMNeXt~(\citeauthor{zhang2023delivering}) & MiT-B4 &59.9 \\ 
        \rowcolor{gray!15} (RGB) \nMethod{} & Ours-B &59.5\\
        \rowcolor{gray!15} (RGBD) \nMethod{} & Ours-L &\highlight{60.3}\\
        \bottomrule
        \end{tabular}\label{tab:MFNet}
  \end{minipage}
  \small
  \hfill
  \begin{minipage}{\linewidth}
   \centering
  \small
  \caption{ KITTI-360 (RGB-L)~(\citeauthor{liao2021kitti360}).
  }\label{tab:KITTI1}
  \vspace{-10pt}
  \renewcommand{\arraystretch}{0.8}
 \begin{tabular}{l|cc}
        \toprule
        \textbf{Method} & \textbf{Backbone} & \textbf{mIoU (\%) } \\
        \midrule\midrule
        HRFuser~(\citeauthor{broedermann2022hrfuser})& HRFormer-T & 48.7 \\
        PMF~(\citeauthor{zhuang2021perception}) & SalsaNext& 54.5 \\
        TokenFusion~(\citeauthor{wang2022multimodal}) & MiT-B2 & 54.6 \\
        TransFuser~(\citeauthor{prakash2021multi}) & RegNetY & 56.6 \\
        CMX~(\citeauthor{zhang2022cmx}) & MiT-B2 & 64.3 \\
        CMNeXt~(\citeauthor{zhang2023delivering}) & MiT-B2 & 65.3\\ 
       \rowcolor{gray!15} (RGB) \nMethod{} & Ours-B & 65.2\\ 
        \rowcolor{gray!15} (RGBD) \nMethod{} & Ours-L & \highlight{66.1}\\ 
        \bottomrule
        \end{tabular}
  \label{tab:KITTI}
  \vspace{-10pt}
  \end{minipage}
\vspace{-10pt}
\end{table}

\section{Experimental Details}
\label{sec:setting}
\subsection{ImageNet Pretraining} 
\label{subsec:setting}
We provide \nMethod{s}' ImageNet-1K pretraining settings in~\tabref{tab:train_detail}.  All \nMethod{} variants use the same settings, except the stochastic depth rate.
Besides, the data augmentation strategies related to color, \eg auto contrast, are only used for RGB images, while other common strategies are simultaneously performed on RGB images and depth maps, \eg random rotation.

\subsection{RGB-D Semantic Segmentation Finetuning}

The finetuning settings for NYUDepth v2 and SUNRGBD datasets are listed in \tabref{tab:fine_set}.
The batch sizes, input sizes, base learning rate, epochs and stochastic depth are different for the two datasets.

\myPara{Datasets.}
Following the common experiment settings of RGB-D semantic segmentation methods~\citep{xie2021segformer,guo2022segnext}, we finetune and evaluate the \nMethod{} on two widely used datasets, \ie NYUDepthv2~\citep{silberman2012nyu_dataset} and SUN-RGBD~\citep{song2015sun_rgbd}.
To be specific, NYUDepthv2~\citep{silberman2012nyu_dataset} contains 1,449 RGB-D samples covering 40 categories, where the resolution of all RGB images and depth maps is unified as $480\times 640$. 
Particularly, 795 image-depth pairs are used to train the RGB-D model, and the remaining 654 are utilized for testing. 
SUN-RGBD~\citep{song2015sun_rgbd} includes 10,335 RGB-D images with $530\times 730$ resolution, where the objects are in 37 categories. 
All samples of this dataset are divided into 5,285 and 5,050 splits for training and testing, respectively.

\myPara{Implementation Details.}
%
%
%
Following SegNext~\cite{guo2022segnext}, we employ Hamburger~\citep{geng2021attention}, a lightweight head, as the decoder to build our RGB-D semantic segmentation network.
%
%
During finetuning, we only adopt two common data augmentation strategies, \ie random horizontal flipping and random scaling (from 0.5 to 1.75).
The training images are cropped and resized to $480\times 640$ and $480\times 480$ respectively for NYU Depthv2 and SUN-RGBD benchmarks. 
%
%
Cross-entropy loss is utilized as the optimization objective.
We use AdamW~\citep{kingma2014adam} as our optimizer with an initial learning rate of 6e-5 and the poly decay schedule. 
Weight decay is set to 1e-2.
During testing, we employ mean Intersection over Union (mIoU), which is averaged across semantic categories, as the primary evaluation metric to measure the segmentation performance. 
Following recent works~\citep{zhang2022cmx,wang2022multimodal,zhang2023delivering}, we adopt multi-scale (MS) flip inference strategies with scales $\left \{0.5, 0.75, 1, 1.25, 1.5\right \}$.
%

\begin{table}[!ht]
\footnotesize
\caption{\textbf{DFormer ImageNet-1K pretraining settings}. All the pretraining experiments are conducted on 8 NVIDIA 3090 GPUs.}
\resizebox{0.99\linewidth}{!}{
\renewcommand{\arraystretch}{1.15}
\begin{tabular}{lcccc}
\multirow{1}{*}{pretraining config} & \nMethod{}-T & \nMethod{}-S & \nMethod{}-B & \nMethod{}-L \\
\toprule
input size& 224$^2$ & 224$^2$ & 224$^2$ & 224$^2$\\
weight init & trunc. normal (0.2) & trunc. normal (0.2) & trunc. normal (0.2) & trunc. normal (0.2) \\
optimizer & AdamW & AdamW & AdamW & AdamW\\
base learning rate & 1e-3 & 1e-3 & 1e-3 & 1e-3\\
weight decay & 0.05 & 0.05& 0.05 & 0.05 \\
optimizer momentum & $\beta_1, \beta_2{=}0.9, 0.999$ & $\beta_1, \beta_2{=}0.9, 0.999$ & $\beta_1, \beta_2{=}0.9, 0.999$ & $\beta_1, \beta_2{=}0.9, 0.999$ \\
batch size & 1024 & 1024 & 1024 & 1024 \\
training epochs & 300 & 300 & 300 & 300 \\
learning rate schedule & cosine decay & cosine decay& cosine decay & cosine decay \\
warmup epochs & 5 & 5 & 5 & 5 \\
warmup schedule & linear & linear & linear & linear \\
layer-wise lr decay& None & None & None & None\\
randaugment & (9, 0.5) & (9, 0.5) & (9, 0.5) & (9, 0.5)\\
mixup & 0.8 & 0.8  & 0.8 & 0.8 \\
cutmix& 1.0 & 1.0& 1.0 & 1.0 \\
random erasing & 0.25 & 0.25& 0.25 & 0.25 \\
label smoothing& 0.1 & 0.1& 0.1 & 0.1 \\
stochastic depth & 0.1 & 0.1 & 0.15 & 0.2 \\
head init scale & None & None& None & None \\
gradient clip & None & None& None & None \\
exp. mov. avg. (EMA)  & None & None& None & None\\
\bottomrule
\end{tabular}}
\label{tab:train_detail}
\end{table}

\begin{table}[!ht]
\footnotesize
\caption{\textbf{DFormer finetuning settings on NYUDepthv2/SUNRGBD}. Multiple stochastic depth rates, input sizes and batch sizes are for NYUDepthv2 and SUNRGBD datasets respectively. All the finetuning experiments for RGB-D semantic segmenations are conducted on 2 NVIDIA 3090 GPUs.}
\resizebox{0.99\linewidth}{!}{
\renewcommand{\arraystretch}{1.15}
\begin{tabular}{lcccc}
\multirow{1}{*}{pretraining config} & \nMethod{}-T & \nMethod{}-S & \nMethod{}-B & \nMethod{}-L \\
\toprule
input size& $480\times 640$ / $480^{2}$ & $480\times 640$ / $480^{2}$ & $480\times 640$ / $480^{2}$ & $480\times 640$ / $480^{2}$\\
optimizer & AdamW & AdamW & AdamW & AdamW\\
base learning rate & 6e-5/8e-5 & 6e-5/8e-5 & 6e-5/8e-5 & 6e-5/8e-5\\
weight decay & 0.01 & 0.01& 0.01 & 0.01 \\
batch size &8/16&8/16&8/16&8/16\\
epochs&500/300&500/300&500/300&500/300\\
optimizer momentum & $\beta_1, \beta_2{=}0.9, 0.999$ & $\beta_1, \beta_2{=}0.9, 0.999$ & $\beta_1, \beta_2{=}0.9, 0.999$ & $\beta_1, \beta_2{=}0.9, 0.999$ \\
training epochs & 300 & 300 & 300 & 300 \\
learning rate schedule & linear decay & linear decay& linear decay & linear decay \\
warmup epochs & 10 & 10 & 10 & 10 \\
warmup schedule & linear & linear & linear & linear \\
layer-wise lr decay& None & None & None & None\\
aux head& None & None& None & None \\
stochastic depth & 0.1/0.1 & 0.1/0.1 & 0.1/0.1 & 0.15/0.2 \\
\bottomrule
\end{tabular}}
\label{tab:fine_set}
\end{table}

\begin{figure}[!ht]
\centering
\vspace{5pt}
\includegraphics[width=0.97\linewidth]{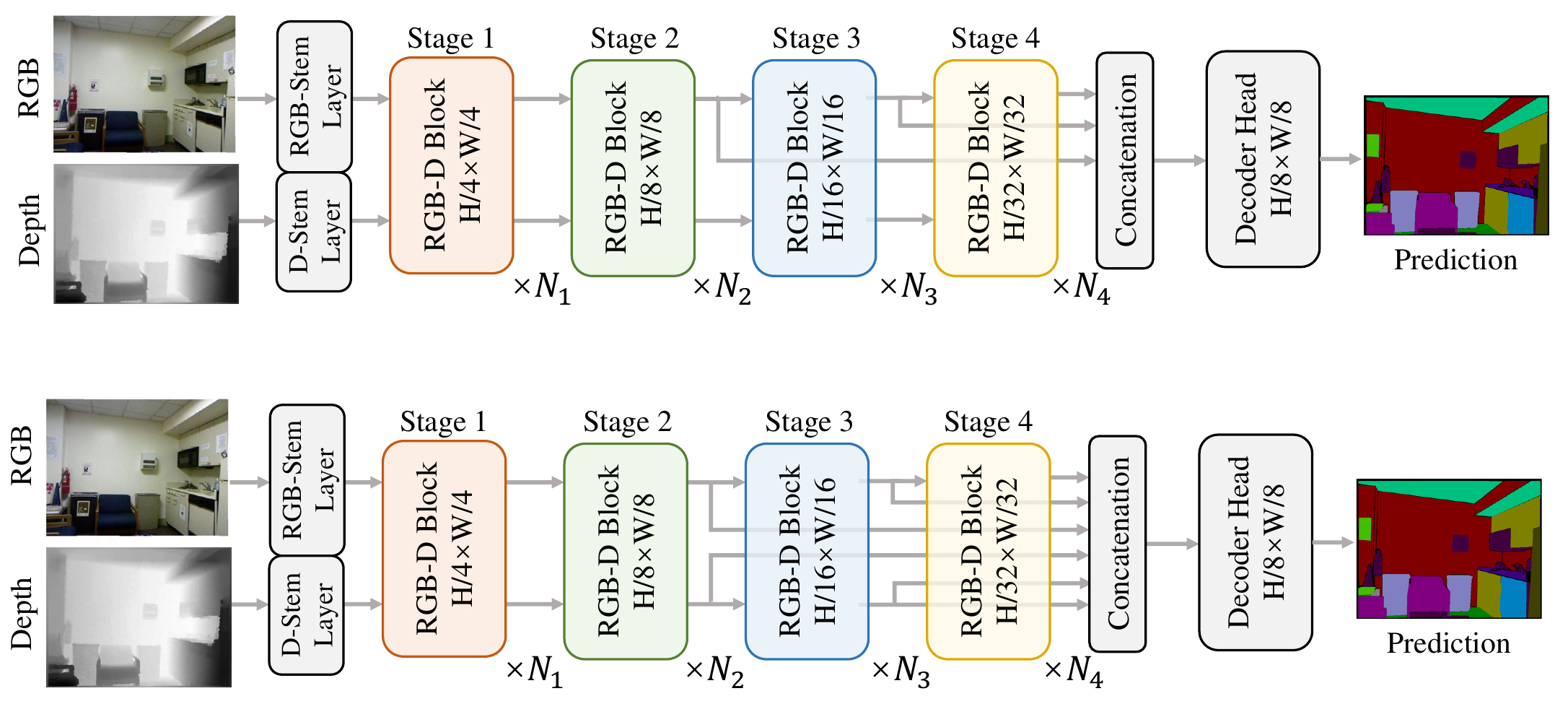}
\vspace{-5pt}
\caption{Detailed illustration for the input features of the decoder head. Top: Only the features in the RGB branch are sent to the decoder. Bottom: The features of both RGB and depth branches are sent to the decoder.
}\label{fig:decoder_in}
\vspace{-1pt}
\end{figure}

\subsection{RGB-D Salient Object Detection}

\myPara{Dataset.}
We finetune and test \nMethod{} on five popular RGB-D salient object detection datasets. 
The finetuning dataset consists of 2,195 samples, where 1,485 are from NJU2K-train~\citep{ju2014depth} and the other 700 samples are from NLPR-train~\citep{peng2014rgbd}.
The model is evaluated on eight datasets, \ie DES~\citep{cheng2014depth} (135 samples), NLPR-test~\citep{peng2014rgbd} (300), NJU2K-test~\citep{ju2014depth} (500), STERE~\citep{niu2012leveraging} (1,000),  SIP~\citep{fan2020rethinking} (929).
The comparison of our method and other \sArt methods is shown in the \tabref{tab:sod}.

\myPara{Implementation Details.}
We set the output channel of Hamburger~\citep{geng2021attention} head to 1, which is further added on the top of our RGB-D backbone to build the RGB-D salient object detection network.
For model finetuning, we adopt the same data augmentation strategies and model optimizer as in SPNet~\citep{zhou2021specificity}.
%
For performance evaluation, we adopt four golden metrics of this task, \ie Structure-measure (S)~\citep{fan2017structure}, mean absolute error (M)~\citep{perazzi2012saliency}, max F-measure (F)~\citep{margolin2014evaluate}, and max E-measure (E)~\citep{fan2018enhanced}. 

\subsection{More details about the decoder input features}
Benefiting from the powerful RGB-D pretraining, the features of the RGB branch can better fuse the information of the two modalities compared to the ones with RGB pretraining.
Thus, our decoder only takes the RGB features $X^{rgb}_{i}$ instead of both $X^{rgb}_{i}$ and $X^{d}_{i}$ as input.
The detailed structures of the two forms are shown in \figref{fig:decoder_in}, as a supplement for \tabref{tab:intersup} in the main paper.
%



\section{More visualization results}\label{sec:vis}
\begin{figure}[!hb]
\centering
\includegraphics[width=0.98\linewidth]{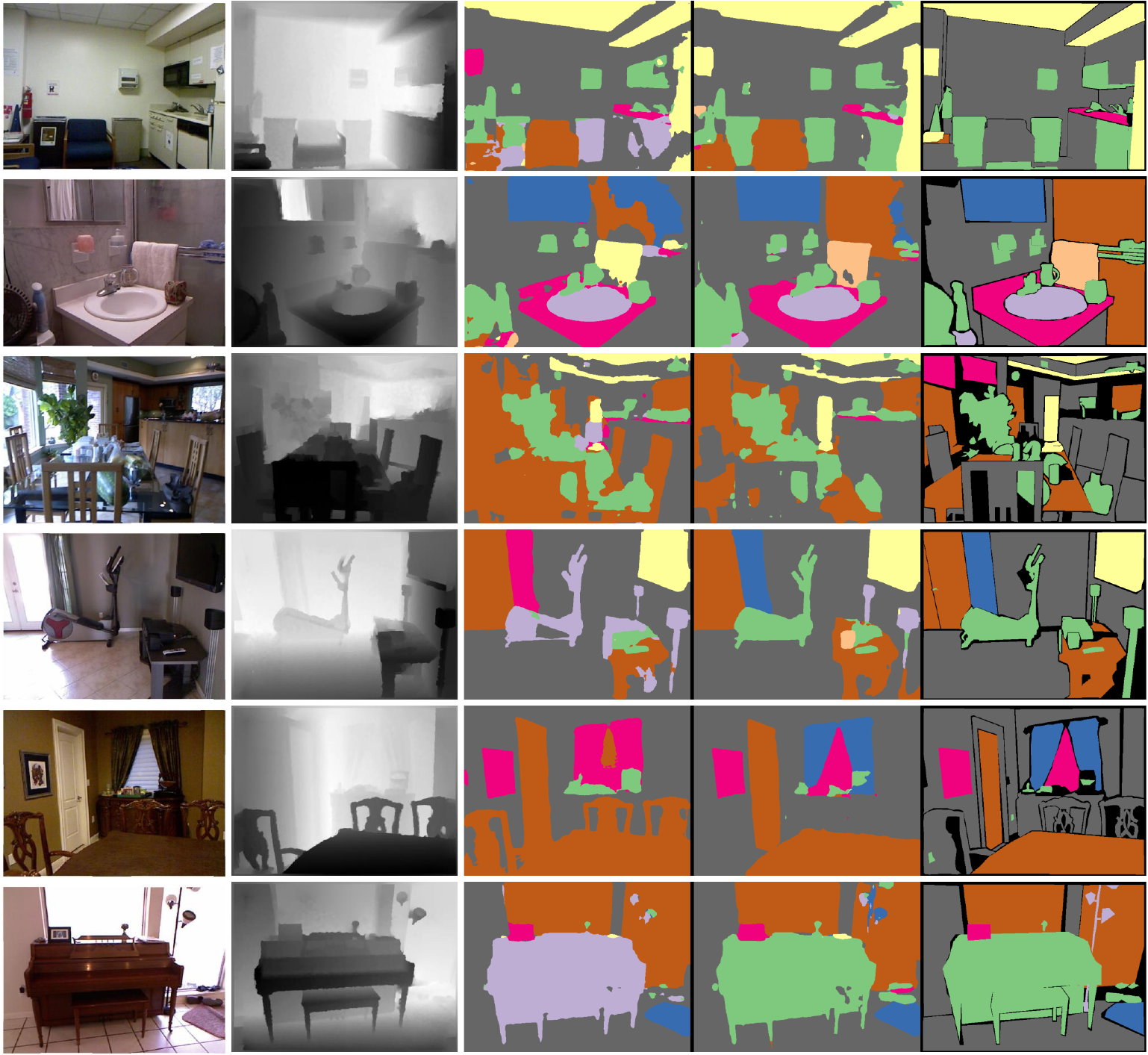}
\vspace{-10pt}
\put (-367, -9){\small Image}
\put (-290, -9){\small Depth}
\put (-209, -9){\small CMNext}
\put (-124, -9){\small Ours}
\put (-50, -9){\small GT}
\caption{\small Qualitative comparison of \nMethod{}-L (Ours) and CMNext (MiT-B4)~\citep{zhang2023delivering} on the NYU Depthv2~\citep{silberman2012nyu_dataset} dataset.}\label{fig:more_vis}
\vspace{-10pt}
\end{figure}

\begin{figure}[!ht]
\centering
\vspace{5pt}
\includegraphics[width=0.97\linewidth]{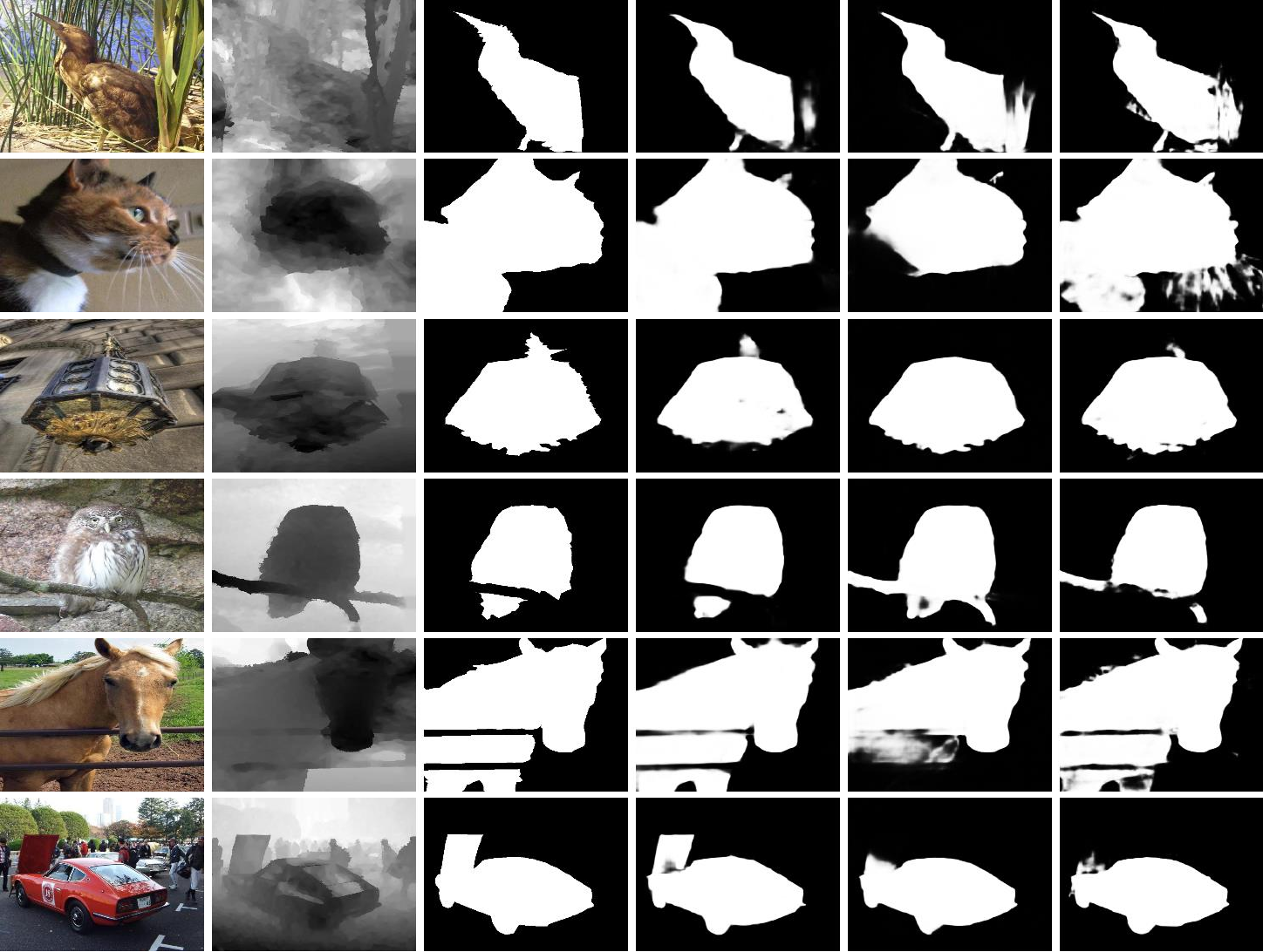}
\vspace{5pt}
\put (-367, -9){\small Image}
\put (-302, -9){\small Depth}
\put (-235, -9){\small GT}
\put (-174, -9){\small Ours}
\put (-117, -9){\small HiDANet}
\put (-45,-9){\small SPNet}
\vspace{-10pt}
\caption{\small Qualitative comparison of \nMethod{}-L (Ours), HiDANet~\citep{wu2023hidanet}, and SPNet~\cite{zhou2021specificity} on the NJU2K dataset.
}\label{fig:sod_vis}
\vspace{-1pt}
\end{figure}

In this section, we provide more visualization results in \figref{fig:more_vis}.
Our \nMethod{} produces higher segmentation accuracy than the current \sArt CMNext (MiT-B4).
Moreover, the visualization comparison on RGB-D salient object detection are shown in \figref{fig:sod_vis}.

\section{Future Work}\label{sec:future}
We argue that there are a number of foreseeable directions for future research:
1) 
%
Applications on a wider range of RGB-D downstream tasks.
Considering that our pretrained \nMethod{} backbones own better RGB-D representation capabilities, it is promising to apply them to more RGB-D tasks, \eg RGB-D face anti-spoofing~\citep{george2021cross}, 3D visual grounding~\citep{liu2021refer}, RGB-D tracking~\citep{yan2021depthtrack}, and human volumetric capture~\citep{yu2021function4d}.
2) 
Extension to other modalities. 
Based on the well-designed framework, it is interesting to produce multi-modal representations by substituting depth with other sensor data in other modalities, such as thermal and lidar data. 
The incorporation of additional modalities can facilitate the development of specialized and powerful encoders to meet the needs of various tasks. 
3) More lightweight models.
Current RGB-D methods are usually too heavy to deploy to mobile devices.
Although \nMethod{} is efficient, there is still a gap and development space in practical applications.
We hope this paper could inspire researchers to develop better RGB-D encoding method, pushing the boundaries of what's feasible in real-world scenarios.

\end{document}